\PassOptionsToPackage{capitalize}{cleveref}
\documentclass{applemlr}

\usepackage{amsmath}
\usepackage{enumerate}
\usepackage{algorithm}
\usepackage{algpseudocode}
\usepackage{amsfonts}
\usepackage{amsthm}
\usepackage{cleveref}
\usepackage{diagbox}
\usepackage{colortbl}
\usepackage{amssymb}
\usepackage{xspace}
\usepackage{wrapfig}
\usepackage{adjustbox}
\usepackage{tabularx}
\usepackage{booktabs}
\usepackage{mathtools}
\usepackage{tikz}
\usepackage{enumitem}
\usepackage{silence}
\usepackage{dsfont}
\usepackage[table]{xcolor}
\usepackage[dvipsnames]{xcolor}
\usepackage{multirow}
\usepackage{makecell}
\usepackage{xfakebold}

\usepackage{amsmath,amsfonts,bm}

\def\eqref#1{equation~\ref{#1}}

\def\1{\bm{1}}

\DeclareMathAlphabet{\mathsfit}{\encodingdefault}{\sfdefault}{m}{sl}
\SetMathAlphabet{\mathsfit}{bold}{\encodingdefault}{\sfdefault}{bx}{n}

\definecolor{textgray}{HTML}{6E6E73}
\usetikzlibrary{positioning, calc}
\usetikzlibrary{decorations.pathmorphing}

\makeatletter
\patchcmd{\wrong@fontshape}{\@gobbletwo}{}{}{}
\makeatother
\numberwithin{equation}{section}
\makeatletter
\AtBeginDocument{
  \urlstyle{sf}
  
}
\makeatother

\definecolor{light}{RGB}{125, 125, 125}
\crefname{tcb@cnt@pbox}{code}{code}
\Crefname{tcb@cnt@pbox}{Code}{Code}
\crefname{assumption}{assumption}{assumption}
\Crefname{assumption}{Assumption}{Assumptions}

\newtcolorbox[auto counter]{pbox}[2][]{
  colback=white,
  title=Code~\thetcbcounter: #2,
  #1,fonttitle=\sffamily,
  fontupper=\sffamily,
  arc=2pt,
  colframe=bgcolor,
  coltitle=fgcolor,
  colbacktitle=bgcolor,
  toptitle=0.25cm,
  bottomtitle=0.125cm
}

\makeatletter
\newcommand\applefootnote[1]{%
  \begingroup
  \renewcommand\thefootnote{}%
  \renewcommand\@makefntext[1]{\noindent##1}%
  \footnote{#1}%
  \addtocounter{footnote}{-1}%
  \endgroup
}
\makeatother

\definecolor{cverbbg}{gray}{0.90}


\newlength{\rowpad}
\newcommand{\padrow}{\rule[-\dimexpr 2.85pt+\rowpad/2\relax]{0pt}{\dimexpr 9.5pt+\rowpad\relax}}
\usepackage{fancyvrb}            
\usepackage{listings}            
\usepackage{pifont}

\newcommand{\meanstd}[2]{%
  \ensuremath{#1{\scriptstyle\,\pm\,#2}}%
}

\newcommand{\bestmeanstd}[2]{%
  \ensuremath{\mathbf{#1}{\scriptstyle\,\pm\,#2}}%
}
\definecolor{cPit}{HTML}{C0392B}    
\definecolor{cRec}{HTML}{1E8449}    
\definecolor{cMiss}{HTML}{B7791F}   
\definecolor{cRule}{HTML}{4A4A4A}
\definecolor{cBg}{HTML}{F5F5F7}

\definecolor{bWall}{HTML}{3D4451}
\definecolor{bFloor}{HTML}{F4F4F2}
\definecolor{bDoor}{HTML}{E0A106}
\definecolor{bObj}{HTML}{B9BDC4}
\definecolor{bTarget}{HTML}{C0392B}
\definecolor{bPath}{HTML}{2E86AB}
\definecolor{bAgent}{HTML}{1E8449}

\newcommand{\vdn}[2]{\ensuremath{#1\,\textcolor{cPit}{\downarrow}\,#2}}
\newcommand{\vup}[2]{\ensuremath{#1\,\textcolor{cRec}{\uparrow}\,#2}}
\newcommand{\vbad}[2]{\ensuremath{#1\,\textcolor{cMiss}{\uparrow}\,#2}}
\newcommand{\vsame}[2]{\ensuremath{#1\,\textcolor{cMiss}{\rightarrow}\,#2}}
\newcommand{\vyes}{\textcolor{cRec}{\ding{51}}}
\newcommand{\vno}{\textcolor{cPit}{\ding{55}}}

\newcommand{\vfield}[1]{%
  \par\smallskip\noindent
  {\scriptsize\sffamily\bfseries\color{cRule}\MakeUppercase{#1}}\par\nobreak\vspace{1.5pt}}

\newtcolorbox{valuebox}[2][]{%
  breakable, enhanced, colback=white, colframe=cRule!80,
  boxrule=0.5pt, arc=1.5pt, left=7pt, right=7pt, top=5pt, bottom=6pt,
  fonttitle=\small, colbacktitle=cBg, coltitle=black,
  title={#2}, #1}

\newtcolorbox{vquote}{%
  enhanced, breakable, colback=cBg, colframe=cBg, boxrule=0pt,
  borderline west={2pt}{0pt}{cRule!35},
  left=8pt, right=6pt, top=4pt, bottom=4pt,
  fontupper=\footnotesize\itshape}

\newenvironment{vtrace}
  {\par\smallskip\centering\footnotesize
   \begin{tabular}{@{}r@{\hspace{7pt}}p{0.52\linewidth}@{\hspace{10pt}}c@{\hspace{10pt}}c@{}}
   \toprule
   \scriptsize Step & \scriptsize Agent action $\rightarrow$ environment response
   & \scriptsize $V^{\mathrm{MC}}$ & \scriptsize $V^{\mathrm{pred}}$\\
   \midrule}
  {\bottomrule\end{tabular}\par\smallskip}

\lstdefinestyle{agent}{%
  language=Python, basicstyle=\ttfamily\scriptsize,
  breaklines=true, columns=fullflexible, keepspaces=true,
  showstringspaces=false, frame=leftline, framesep=6pt,
  rulecolor=\color{cRule!25}, xleftmargin=7pt,
  aboveskip=3pt, belowskip=2pt,
  keywordstyle=\color{cRule}, stringstyle=\color{cRule!70},
  commentstyle=\itshape\color{gray},
  moredelim=**[is][\color{cPit}\bfseries]{@}{@}}
\newlength{\bcell}

\tikzset{
  bwall/.style       = {fill=bWall},
  bfloor/.style      = {fill=bFloor},
  bdoor/.style       = {fill=bDoor},
  bobj/.style        = {fill=bObj},
  btarget/.style     = {fill=bTarget},
  btargetring/.style = {draw=bTarget,line width=0.5pt},
  bagent/.style      = {fill=bAgent},
  bpathcase/.style   = {draw=white,line width=3.2pt,
                        line cap=round,line join=round},
  bpath/.style       = {draw=bPath,line width=1.5pt,opacity=0.85,
                        line cap=round,line join=round},
}

\newcommand{\fullmethodrl}{LLM Value Function RL\xspace}
\newcommand{\methodrl}{LVRL\xspace}
\newcommand{\fullmethodopd}{LLM Value Function OPD\xspace}
\newcommand{\methodopd}{LVOPD\xspace}

\newcommand{\traj}{\tau}
\newcommand{\Egroup}{\mathcal{G}}
\newcommand{\Vllm}{V_{\phi}}

\newcommand{\pist}{\pi_{\theta}}
\newcommand{\pite}{\pi_{E}}

\newcommand{\kl}{\mathrm{KL}}

\usepackage[textwidth=18mm]{todonotes}
\usepackage{bbm}

\title{LLM Value Functions for Credit Assignment in Agentic RL}

\author[*]{Kaijie Zhu}
\author{Weixiang Yan}
\author{Omar Attia}
\author{Abbas Kazerouni}
\author{Alexander Toshev}
\author{Andrew Szot}

\affiliation{Apple}

\abstract{Reinforcement learning (RL) for large language model (LLM) agents remains challenging in long-horizon tasks with sparse outcome rewards, where assigning credit to individual actions is difficult. 
Methods such as GRPO assign the same trajectory-level advantage to all actions, while critic-based methods require learning an additional value function that can be costly and unstable. 
We introduce LLM Value Function RL (LVRL), which uses the pretrained knowledge and in-context learning ability of an LLM to estimate policy-aware state values without finetuning a separate critic. 
LVRL conditions the value model on group context and calibrates its predictions using known values at the initial and terminal states, enabling step-level credit assignment from estimated value differences. 
We further extend this idea to learning from an expert policy through on-policy distillation (OPD), introducing LLM Value Function OPD (LVOPD), which focuses teacher supervision on actions associated with decreases in estimated value. 
We evaluate our approach across diverse agentic benchmarks: AppWorld, BabyAI, and SWE-bench Verified.
Across policy scales (4B to 32B), \methodrl consistently and substantially outperforms both GRPO and PPO. 
On AppWorld and SWE-bench Verified, \methodrl yields absolute improvements of up to $13.6$ and $8.6$ percentage points over GRPO. 
Notably, on tasks requiring extended exploration like BabyAI, \methodrl raises success from near $0\%$ to $75.0\%$. 
Furthermore, \methodopd yields consistent improvements over standard distillation baselines on AppWorld. 
Finally, a Monte-Carlo analysis shows that the frozen critic identifies most semantic pitfalls and recoveries, indicating that \methodrl extracts meaningful, policy-aware step-level credit from sparse terminal outcomes.
}

\metadata[Correspondence]{\sffamily Andrew Szot: \url{a_szot@apple.com}; Kaijie Zhu: \url{kaijiezhu@ucsb.edu}}
\date{\sffamily\today}

\begin{document}

\maketitle

{\renewcommand\thefootnote{*}\footnotetext{\sffamily Work done as part of internship at Apple.}}
\applefootnote{ \textcolor{textgray}{\sffamily Apple and the Apple logo are trademarks of Apple Inc., registered in the U.S. and other countries and regions.}}

\section{Introduction}
\label{sec:intro}

Language models are increasingly deployed as agents that reason and act over many steps to solve complex tasks~\citep{yao2022react,yang2024sweagent,jimenez2024swe,trivedi2024appworld}.
Reinforcement learning allows these agents to improve from experience, but environment rewards are typically sparse and outcome-based in the form of a single binary signal at the end of the episode indicating success or failure.
How to distribute this single bit of feedback across individual actions is the classical problem of temporal credit assignment~\citep{sutton1998reinforcement}.

Current policy optimization methods face a trade-off between coarse trajectory-level credit and the high cost of learning a value function.
Outcome-only methods like GRPO~\citep{shao2024deepseekmath} estimate a group-relative baseline and assign the same scalar advantage to every action in a trajectory.
This uniform signal fails to distinguish individual contributions as mistakes in successful trajectories receive positive credit, while useful actions in failed trajectories receive negative credit.
Alternatively, methods like PPO~\citep{schulman2017ppo} learn a value function to estimate step-level advantages.
However, training a critic adds computational overhead and struggles to accurately evaluate long interaction histories from sparse rewards~\citep{kazemnejad2025vineppo}.
While recent work explores alternative step-level credit mechanisms, such as process reward models~\citep{lightman2024let}, Monte-Carlo continuations~\citep{kazemnejad2025vineppo}, or outcome-conditioned heuristics~\citep{tan2026hindsight}, they often require human annotations, extra rollout budgets, or rely on proxy metrics.

To overcome these limitations, we introduce \fullmethodrl (\methodrl), a method that extracts dense, per-step credit using a strong, frozen LLM as a value critic.
By evaluating completed trajectories in hindsight, the frozen LLM critic observes downstream consequences to accurately assess earlier actions, such as ignored tool errors or successful recoveries.
Because value estimation depends on the acting policy, we make the frozen LLM \emph{policy-aware} through in-context learning.
Specifically, we condition the estimator on summaries of other rollouts from the same task, along with their empirical success rate and final rewards.
This dynamically updates the critic's context with the current policy's behavior and task performance, without requiring any parameter updates.
We then convert the predicted values into step-level advantages via consecutive value differences~\citep{ng1999shaping}.
Crucially, by anchoring the initial state to the group success rate and the terminal state to the observed final reward, we guarantee that these step-level advantages sum exactly to the trajectory's overall group-relative return.
The frozen critic thus redistributes sparse outcome rewards across individual actions.

We further extend this credit assignment technique to on-policy distillation (OPD), where a stronger teacher provides token-level supervision~\citep{agarwal2024policy,gu2024minillm}.
Since uniform distillation penalizes large KL divergence everywhere, it can suppress alternative yet successful strategies discovered by the student.
To align distillation with the capabilities of the student, \methodopd introduces \emph{value-guided gating}. 
By restricting teacher supervision to actions that decrease the estimated value, we selectively correct execution errors while preserving the student's independent progress.

We train and evaluate \methodrl and \methodopd on AppWorld, BabyAI, and SWE-bench Verified across policy scales ranging from 4B to 32B parameters.
In \methodrl, we utilize Claude Opus 4.6 as the frozen LLM value critic.
\methodrl consistently improves final generalization over outcome-only supervision (GRPO) and learned value functions (PPO).
For instance, on Qwen3-8B, \methodrl achieves $26.3\%$ success on AppWorld and $20.2\%$ on SWE-bench Verified, compared to $19.8\%$ and $12.0\%$ for GRPO, respectively.
Notably, \methodrl is particularly critical in environments requiring extended exploration: on BabyAI, it reaches $65.5\%$ success on out-of-distribution scenarios, whereas baseline methods fail entirely.
Furthermore, \methodopd yields consistent improvements over standard distillation baselines on AppWorld, reaching a $20.2\%$ success rate with Qwen3-4B and improving pure OPD by $3.2$ percentage points.

\begin{figure}[t]
\centering
\includegraphics[width=0.95\textwidth]{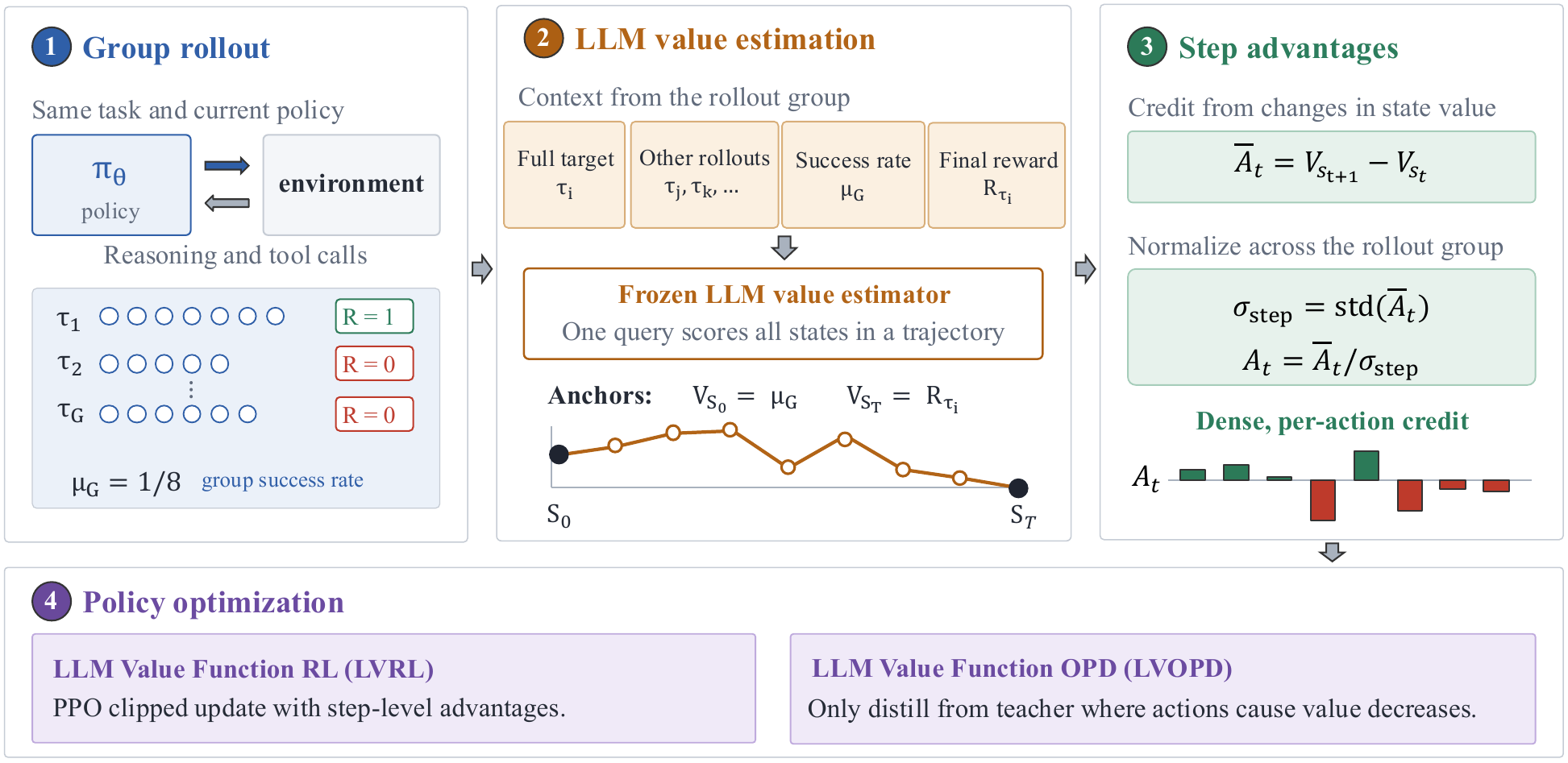}
\caption{
  Overview of the proposed framework.
  (\textbf{1}) \textbf{Group rollout:} We sample a group of trajectories from the current policy and use their sparse outcome rewards to compute the empirical group success rate $\mu_G$.
  (\textbf{2}) \textbf{LLM value estimation:} A frozen LLM critic scores all trajectory states, conditioned on group context. Estimates are anchored to empirical outcomes: the initial-state value to the group success rate and the terminal-state value to the actual reward.
  (\textbf{3}) \textbf{Step advantages:} We compute unnormalized step advantages as consecutive value differences ($\overline{A}_t = \Vllm(s_{t+1}) - \Vllm(s_t)$) and normalize them across the group, yielding a dense, per-action credit signal.
  (\textbf{4}) \textbf{Policy optimization:} These step-level advantages drive standard PPO updates in \methodrl, or act as a gating mechanism in \methodopd to restrict teacher supervision to actions that decrease the estimated value.
}
\label{fig:method}
\vspace{-.15in}
\end{figure}

Finally, we analyze the accuracy of the frozen LLM critic's value predictions. 
Using Monte-Carlo rollouts as a ground-truth probe, we demonstrate that predicted value changes correctly identify $72$\% of critical pitfalls and $66$\% of recoveries.
Furthermore, our ablations highlight two insights: in-context policy calibration is important for maximizing improvements, and the method's overall efficacy is bounded by the capability of the chosen LLM critic.
In summary, we present a unified framework for leveraging a frozen, policy-aware LLM to provide per-action credit assignment to guide RL and on-policy distillation, all without critic updates.

\section{Related Work}
\label{sec:related}


\textbf{Step-level credit assignment for LLM agents.}
Classical actor--critic methods address temporal credit assignment by learning a parameterized value function~\citep{schulman2015high,schulman2017ppo}, an approach extended to multi-turn language agents by ArCHer~\citep{zhou2024archer} and GenAC~\citep{shan2026bringing}.
To avoid the overhead of critic optimization, recent methods explore alternative credit mechanisms.
The GRPO family of algorithms~\citep{shao2024deepseekmath, liu2025drgrpo, yu2025dapo} relies on group-relative trajectory returns, uniformly assigning the same advantage to all actions within a rollout.
Another line of work relies on extensive Monte-Carlo sampling; VinePPO~\citep{kazemnejad2025vineppo} explicitly rolls out continuations from intermediate states to estimate state values, whereas GiGPO~\citep{feng2025groupingroup} and SALT~\citep{li2026salt} achieve a similar effect implicitly by aggregating returns over shared states across multiple trajectories.

Other approaches rely on auxiliary models or proxy signals.
Process reward models are trained with intermediate supervision~\citep{lightman2024let,zhou2025sweet,wang2025spa,xi2026agentprm,wu2026repirl}, while implicit estimators derive step- or token-level credit from hindsight judgments of the policy LLM~\citep{tan2026hindsight}, target-answer likelihoods~\citep{wang2025information,tao2026trace,oh2026neglected}, or internal signals of the policy, such as the discounted KL divergence of subsequent tokens~\citep{ma2026fipo}, policy-to-reference log-ratios~\citep{kang2026vimpo}, and attention patterns~\citep{li2025attention}.
Since these implicit estimators are tailored for single-turn reasoning with static answers, adapting them to multi-turn agentic tasks would require non-trivial modifications to handle dynamic environment feedback, complicating direct comparisons.
Conceptually related to our approach, GVL~\citep{ma2025vision} uses a frozen vision-language model to estimate task progress from robot videos via in-context learning.
However, GVL produces policy-agnostic progress scores.
\methodrl conditions the LLM on rollouts from the current policy and explicitly anchors its predictions to observed outcomes. 
This formulation ensures that the estimated values are both dynamically policy-aware and consistent with actual environment returns.

\textbf{On-policy distillation.}
Knowledge distillation transfers capabilities from a stronger teacher to a student model~\citep{hinton2015distilling}.
For generative tasks, on-policy distillation (OPD) mitigates exposure bias by evaluating the teacher on prefixes generated by the student~\citep{agarwal2024policy}, often optimizing a reverse-KL objective~\citep{gu2024minillm}.
However, standard token-level distillation is overly rigid, penalizing deviation from the teacher even when the student discovers a valid alternative strategy.
Recent methods adapt teacher-guided updates using token- or step-level likelihood differences~\citep{jia2026asymmetric,zhong2026sod}, verifier rewards combined with teacher signals~\citep{akhondzadeh2026reward,yang2026self,xu2026sg,lin2026policy}, or environment-informed teacher signals~\citep{zhou2026sage,li2026what}.
SAGE-OPD explicitly allows valid alternative actions to skip supervision~\citep{zhou2026sage}, while FutureBridge-OPD validates candidate teacher interventions through student continuations scored using teacher-relative preferences~\citep{chen2026look}.
These approaches typically rely on specialized heuristics or complex combinations of environment and teacher signals to define their gating rules.
Our approach, \methodopd, offers a unified and streamlined alternative: it weights reverse-KL distillation by the negative part of consecutive value differences estimated by the frozen LLM critic.
Actions with non-negative predicted value changes therefore receive no distillation loss, preserving the student's valid alternative strategies even when they differ from the teacher's preferred behavior.

\section{Method}
\label{sec:method}

\subsection{Preliminaries}
\label{sec:prelim}

We model agentic tasks as finite-horizon MDPs over interaction histories.
At environment step $t$, the state $s_t$ contains the goal $g$ and the history of observations and actions preceding action $a_t$.
The LLM policy samples $a_t\sim\pi_\theta(\cdot\mid s_t)$ as a sequence of tokens.
A trajectory $\tau=(s_0,a_0,\ldots,s_{T-1},a_{T-1},s_T,r_T)$ receives only a binary terminal reward $R(\tau)=r_T\in\{0,1\}$, indicating task success.
With discount factor $\gamma=1$, the objective is to maximize $\mathbb{E}_{\tau\sim\pi_\theta}[R(\tau)]$.
For $t<T$, the value function $V^\pi(s_t)=\mathbb{E}_{\tau\sim\pi}[R(\tau)\mid s_t]$ represents the probability of eventual task success.
Throughout, $t$ indexes environment steps rather than individual tokens.

PPO~\citep{schulman2017ppo} learns a value estimator $\hat{V}_\psi(s_t)$ and computes advantages $\hat{A}_t$ from observed rewards and value estimates using Generalized Advantage Estimation (GAE)~\citep{schulman2015high}.
For trajectories sampled from $\pi_{\theta_{\mathrm{old}}}$, let $r_t(\theta)=\pi_\theta(a_t\mid s_t)/\pi_{\theta_{\mathrm{old}}}(a_t\mid s_t)$ denote the likelihood ratio.
PPO maximizes the following clipped surrogate where $\epsilon>0$ is the clipping coefficient and $\widehat{\mathbb{E}}_t$ denotes an empirical average over sampled steps.
\begin{equation}
\mathcal{J}_{\mathrm{PPO}}(\theta) =
\widehat{\mathbb{E}}_t
\left[
\min\left(
r_t(\theta)\hat{A}_t,\,
\operatorname{clip}\left(r_t(\theta),1-\epsilon,1+\epsilon\right)\hat{A}_t
\right)
\right],
\label{eq:ppo}
\end{equation}
To avoid learning a critic, GRPO~\citep{shao2024deepseekmath} computes a baseline from a group of $G$ trajectories $\{\tau_1, \ldots, \tau_G\}$ sampled from $\pi_{\theta_{\mathrm{old}}}$ for the same task.
It assigns a uniform advantage $\hat{A}_i=(R(\tau_i)-\mu_G)/\sigma_G$ to \emph{every} action in $\tau_i$, where $\mu_G$ and $\sigma_G$ are the group mean and standard deviation of the rewards.
In multi-step tasks, this uniform credit inadvertently reinforces mistakes in successful trajectories and penalizes useful actions in failed trajectories.

\subsection{\fullmethodrl}
\label{sec:value-rl}

\fullmethodrl (\methodrl) uses a frozen LLM to estimate state values from completed trajectories and converts these estimates into step-level advantages without requiring critic updates.
The overall pipeline is illustrated in \cref{fig:method}.

The core of our approach lies in utilizing the frozen LLM, denoted as \(\Vllm\), to generate step-level value estimates.
For each task, we sample a group of trajectories $\Egroup=\{\traj_1, \ldots, \traj_G\}$ from the current policy and compute their mean return $\mu_G$. 
For each target trajectory $\traj_i$, we randomly select $k$ other trajectories from $\Egroup$ as context. 
A single query then jointly estimates all interior state values as $\Vllm\!\left(s_{i,t}\mid\traj_i,\{S(\traj_j)\}_{j\in\mathcal{C}_i},\mu_G\right)$, where $\mathcal{C}_i\subseteq\{1,\ldots,G\}\setminus \{i\}$ with $|\mathcal{C}_i|=k$ indexes the context trajectories.
To reduce context length, the summarization function $S(\cdot)$ condenses each context rollout into its essential action--observation history (see \cref{app:llm-prompts}).
We anchor the endpoints to empirical data: the initial state to the group success rate ($\Vllm(s_{i,0})=\mu_G$) and the terminal state to the observed final reward ($\Vllm(s_{i,T_i})=R(\traj_i)$).

Inspired by potential-based shaping~\citep{ng1999shaping}, we define the unnormalized step-level advantage as \(\overline{A}_{i,t}=\Vllm(s_{i,t+1})-\Vllm(s_{i,t})\). 
Since intermediate environment rewards are zero, setting \(\gamma=1\) makes this formulation mathematically equivalent to Generalized Advantage Estimation (GAE) at \(\lambda=0\).
However, $V_\phi$ is conditioned on the completed trajectory and its outcome, so it is a hindsight estimate rather than a true state-value function.
Let $\sigma_{\mathrm{step}}$ denote the standard deviation of these unnormalized advantages across all valid environment steps in the entire rollout group $\Egroup$, computed over the set $\{\overline{A}_{j,t} \mid j \in \{1,\ldots,G\}, 0 \le t < T_j\}$. 
We compute the final normalized step advantage as $A_{i,t}=\overline{A}_{i,t}/\sigma_{\mathrm{step}}$.
\methodrl then replaces the learned advantage $\hat{A}_t$ in \cref{eq:ppo} with our normalized step-level advantage $A_{i,t}$. 
While $A_{i,t}$ is computed at the environment-step level, the policy generates each action $a_{i,t}$ as a sequence of tokens. 
Therefore, all tokens within $a_{i,t}$ inherit the exact same $A_{i,t}$, whereas the likelihood ratio and clipping are evaluated per-token. 
The final policy loss $\mathcal{L}_{\mathrm{LVRL}}(\theta; \mathcal{B})$ is the negative empirical mean of this token-level surrogate across the minibatch.

The endpoint anchors guarantee that the unnormalized advantages telescope: $\sum_{t=0}^{T_i-1}\overline{A}_{i,t}=R(\traj_i)-\mu_G$, matching the numerator of the GRPO advantage. 
The normalized advantages satisfy $\sum_{t=0}^{T_i-1}A_{i,t}=(R(\traj_i)-\mu_G)/\sigma_{\mathrm{step}}$. 
The estimator thus redistributes an outcome-grounded, trajectory-level signal across environment steps while preserving its unnormalized sum. 
Note that process reward models (PRMs) inject \emph{learned additive reward} into the RL objective, allowing the policy to exploit estimation errors. 
For example, the policy can exploit these errors to accumulate a positive total reward on a trajectory that fails.
Because the per-step advantages obtained by \methodrl sum to the empirical outcome advantage, the frozen critic cannot inject new reward; it only redistributes it. 
Estimator error is strictly confined to \emph{within-trajectory} credit assignment and cannot flip the sign of a trajectory's overall advantage.


\subsection{\fullmethodopd}
\label{sec:value-opd}

\methodrl uses the step-level advantage to determine how sampled behavior contributes to the policy update.
When a stronger teacher model with accessible token distributions is available, this same advantage can serve a complementary role: determining \emph{where} teacher supervision should be applied.
We refer to this joint setting as~\emph{\fullmethodopd}~(\methodopd).

Standard OPD trains the student on its own generated prefixes by minimizing the divergence against token-level targets from a frozen teacher $\pite$.
However, applying this supervision uniformly throughout a trajectory is problematic.
A large KL divergence does not necessarily imply that the student's behavior is incorrect; the student might be employing a different, yet effective, reasoning strategy or surface form (e.g., exploring a software repository in a different but valid order).
Forcing the student to match the teacher everywhere can suppress alternative behaviors that contribute to task success.

The step-level advantage from \cref{sec:value-rl} provides a principled mechanism to disentangle task failure from benign divergence.
When $A_{i,t} \ge 0$, the sampled action is estimated to contribute positively or neutrally to the outcome. 
In this scenario, distillation is either redundant (if the student and teacher align) or potentially counterproductive (if they diverge), as uniform supervision risks penalizing the student's valid alternative strategies. 
Conversely, when $A_{i,t} < 0$, the action reduces the estimated task value. 
On these steps, the RL policy gradient merely pushes the policy \emph{away} from the mistake, whereas distillation supplies a stronger reference distribution to guide the student toward a correct recovery path.
Consequently, teacher supervision is most critical on steps with negative advantages, motivating our selective gating mechanism.

We implement this value-guided distillation by using the step-level advantage to gate the reverse-KL objective.
For trajectory $\traj_i$, let $a_{i,t}=(a_{i,t,1},\ldots,a_{i,t,L_{i,t}})$ denote the action taken at step $t$ with length $L_{i,t}$, and let $h_{i,t,u}=(s_{i,t},a_{i,t,<u})$ denote the prefix preceding token $a_{i,t,u}$.
We define the gating function as $g(A_{i,t}) = \max(-A_{i,t}, 0)$.
This gate activates only on actions that decrease the estimated task value, dynamically scaling the distillation weight by the magnitude of the predicted drop. 
The resulting value-guided distillation loss is:
\begin{equation}
\mathcal{L}_{\mathrm{VD}}(\theta;\mathcal{B})
=
\frac{1}{N_{\mathcal{B}}}
\sum_{\traj_i\in\mathcal{B}}
\sum_{t=0}^{T_i-1}
\sum_{u=1}^{L_{i,t}}
g\!\left(A_{i,t}\right)
\kl\!\left(
\pist(\cdot\mid h_{i,t,u})
\,\|\,
\pite(\cdot\mid h_{i,t,u})
\right),
\label{eq:value-distill}
\end{equation}
where $N_{\mathcal{B}} = \sum_{\traj_i\in\mathcal{B}} \sum_{t=0}^{T_i-1} L_{i,t}$ is the total number of valid action tokens in the minibatch $\mathcal{B}$.
For computational efficiency, we implement this KL divergence using a renormalized top-$k$ approximation over the vocabulary; details are given in \cref{app:lvopd_training_detials}.
Crucially, while the gate $g(A_{i,t})$ is defined at the step level and shared across all tokens of action $a_{i,t}$, the KL divergence is computed over the vocabulary distribution at each token position.
\methodopd jointly optimizes the clipped policy surrogate and this gated distillation loss: 
$
\mathcal{L}_{\mathrm{LVOPD}}(\theta;\mathcal{B}) = \mathcal{L}_{\mathrm{LVRL}}(\theta;\mathcal{B}) + \beta\, \mathcal{L}_{\mathrm{VD}}(\theta;\mathcal{B}),
$
where $\beta\geq0$ controls the distillation strength.

\section{Experiments}
\label{sec:exp}

\subsection{Setup}
\label{sec:exp:setup}

\textbf{Benchmark environments.}
We train and evaluate on three multi-step agent environments spanning distinct capabilities: digital tool use, spatial navigation, and repository-level software engineering.
\textbf{AppWorld} \citep{trivedi2024appworld} requires an agent to complete everyday digital tasks by writing and executing code against a suite of simulated applications.
We train on the $90$ tasks of the AppWorld training tasks.
\textbf{BabyAI}~\citep{chevalier2018babyai} provides text navigation and interaction tasks in a grid world. 
To rigorously test sustained multi-step decision-making, we use a modified Pickup benchmark arranged in a $2 \times 3$ grid of six rooms.
This topology requires the agent to navigate across rooms and open doors to pick up a target object.
We train on $285$ grid configurations with randomized object positions and target door distances in $\{1,2,3\}$.
The policy receives textual egocentric observations and emits one low-level action per step, such as move forward or turn left.
\textbf{R2E-Gym}~\citep{jain2025r2e} provides repository-level software-engineering tasks in which the agent interacts with the codebase through the four tools of the R2E-Gym scaffold (\texttt{search}, \texttt{file\_editor}, \texttt{execute\_bash}, and \texttt{finish}).
We use $3,200$ R2E-Gym tasks and the R2E-Gym agent scaffold for training.
All three environments provide sparse outcome rewards, with task success determined at the end of each trajectory.
Full details about these training environments and the following evaluation environments are in \cref{app:sec1_env}.

\textbf{Training protocol and baselines.}
We fine-tune Qwen3-4B and Qwen3-8B on AppWorld and BabyAI, and Qwen3-8B and Qwen3-32B on R2E-Gym~\citep{qwen3}.
All policies operate in non-thinking mode to isolate the effect of RL optimization.
We run three training seeds on AppWorld and BabyAI and one seed on R2E-Gym due to the higher computational cost of this environment.
All methods use a group size of $G=8$. Full training details are in \cref{app:training_details}.

For \methodrl, we compare against GRPO, which estimates advantages without learning a value function, and PPO, which learns a parameterized value function for advantage estimation.
We use Claude Opus 4.6~\citep{opus46} as the frozen LLM value critic with three in-context examples.
The estimated API costs per trajectory for each environment are summarized in \cref{app:api_cost}.

For \methodopd, we mainly focus our evaluation on AppWorld, comparing against two distillation baselines to isolate the impact of step-level gating.
\emph{Pure OPD} applies teacher supervision uniformly across all generated tokens. 
\emph{Trajectory-gated OPD} replaces both components of LVOPD with their trajectory-level counterparts: it optimizes the GRPO objective and gates the same reverse-KL loss with the GRPO advantage. 
Consequently, it applies a single supervision weight to every step of a rollout.
For our \methodopd implementation, Qwen3-235B-A22B-Thinking-2507-FP8 (Qwen3-235B for brevity) serves as both the distillation teacher and the frozen LLM value critic, with the distillation strength set to \(\beta=1\).

\textbf{Evaluation protocol.}
We evaluate AppWorld policies on the Test Challenge split and R2E-Gym policies on SWE-bench Verified~\citep{jimenez2024swe}.
For BabyAI, we evaluate on $300$ in-distribution layouts (door distances $\in \{1,2,3\}$) and $200$ out-of-distribution layouts (distances $\in \{4,5\}$).
For all environments, we evaluate the final training checkpoint in three independent runs.
For AppWorld and BabyAI, we report the mean and standard deviation across the three training seeds.
For R2E-Gym, we report the mean and standard deviation across the three evaluation runs of the single seed.

\subsection{Does \methodrl Improve Long-Horizon Agent Learning?}
\label{sec:exp:rl}

We evaluate whether \methodrl translates to superior generalization on held-out tasks. 
As shown in \cref{tab:main-rl}, \methodrl consistently outperforms GRPO and PPO across all domains and model scales. 
On AppWorld, \methodrl achieves a success rate of $24.3\%$ with Qwen3-4B and $26.3\%$ with Qwen3-8B. 
This substantially exceeds the strongest baselines at each scale, which achieve only $14.8\%$ (PPO) and $19.8\%$ (GRPO), respectively. 
This performance gap is sustained in the repository-level SWE-bench Verified environment. 
For Qwen3-8B, \methodrl reaches $20.2\%$, substantially exceeding GRPO ($12.0\%$) and PPO ($9.0\%$). 
Furthermore, scaling the policy to Qwen3-32B increases \methodrl's performance to $30.9\%$, outperforming PPO's $24.9\%$.
This demonstrates that \methodrl remains highly effective even for large-capacity models.
Beyond final performance, \methodrl significantly improves overall learning efficiency. 
As illustrated in \cref{app:fig-rl-curve}, \methodrl establishes an early lead in training reward and converges much faster across these environments. 
In contrast, GRPO struggles with slower convergence, and PPO requires substantially more environment interactions to learn a useful critic. 
This suggests that our step-level value guidance improves sample efficiency.

The advantage of step-level credit assignment is most pronounced in the BabyAI setting. 
Because the base models' success rates are near $0\%$, initial rollouts yield almost exclusively zero rewards. 
This creates a severe cold-start bottleneck: GRPO advantages vanish without reward variance, and PPO's learned critic collapses to predicting zero everywhere. 
Consequently, both baselines fail to extract meaningful learning signals, scoring near $0\%$ on both in-distribution (ID) and out-of-distribution (OOD) layouts.
For \methodrl, when all trajectories in a group fail, both endpoint anchors are zero, so the unnormalized advantages sum to zero for every trajectory. 
Nevertheless, interior estimates can vary, producing positive and negative step advantages.
This allows \methodrl to use the critic's inherent semantic capacity to recognize and reward intermediate progress even when a trajectory ultimately fails.
By providing this step-level signal, \methodrl sidesteps the cold-start problem, allowing Qwen3-8B to achieve $75.0\%$ ID success and $65.5\%$ OOD success.

To understand these gains qualitatively, we analyze the step-level credit assigned by \methodrl across environments. 
Unlike GRPO, whose relative advantages vanish entirely if all rollouts in a group fail, the frozen LLM critic in \methodrl can still isolate useful intermediate progress. 
For instance, it assigns positive value differences to constructive steps, such as recovering from API parameter errors in AppWorld, opening a door to reveal a target in BabyAI, or successfully reproducing a bug in R2E-Gym, even when these trajectories ultimately fail. 
Conversely, it assigns negative differences to local errors within successful trajectories, such as incorrect file edits or API calls. 
By explicitly rewarding intermediate semantic progress and penalizing localized mistakes, \methodrl avoids GRPO's uniform trajectory-level credit while circumventing PPO's dependence on learning a critic from sparse returns. 
Detailed training trajectory examples illustrating these value changes are shown in \cref{app:training_traj_examples}.

However, this dense, zero-shot supervision introduces a computational trade-off. 
As detailed in \cref{app:api_cost}, querying Claude Opus 4.6 for trajectory evaluation incurs an inference cost of approximately $9$k to $20$k input tokens and $0.6$k to $1.9$k output tokens per target trajectory, depending on the environment context length. 
Importantly, our framework is not tied to a single proprietary API. 
As shown in our ablations (\cref{sec:ablation}), alternative strong models like GLM-5.2 can also yield competitive policy improvements. 
Distilling this signal into a smaller, more efficient local critic is a promising direction for reducing this cost.

\begin{table}[t]
\centering
\caption{
\methodrl evaluation performance across three environments.
BabyAI results are reported as ID / OOD success rates.
Uncertainties denote standard deviations.
Bold indicates the best mean for each environment and policy size.
The frozen LLM value critic is Claude Opus 4.6.
}
\label{tab:main-rl}

\setlength{\tabcolsep}{2.5pt}
\renewcommand{\arraystretch}{1}

\begin{adjustbox}{max width=\linewidth}
\begin{tabular}{@{}lcccccc@{}}
\toprule
& \multicolumn{2}{c}{AppWorld}
& \multicolumn{2}{c}{BabyAI}
& \multicolumn{2}{c}{SWE-bench Verified} \\
\cmidrule(lr){2-3}
\cmidrule(lr){4-5}
\cmidrule(lr){6-7}
Method
& Qwen3-4B
& Qwen3-8B
& Qwen3-4B
& Qwen3-8B
& Qwen3-8B
& Qwen3-32B \\
\midrule
Base
& \meanstd{4.2}{0.4}
& \meanstd{10.9}{1.0}
& \meanstd{0.0}{0.0}\,/\,\meanstd{0.0}{0.0}
& \meanstd{0.7}{0.0}\,/\,\meanstd{0.0}{0.0}
& \meanstd{2.9}{0.2}
& \meanstd{20.2}{1.2} \\

\midrule
+GRPO
& \meanstd{10.7}{2.8}
& \meanstd{19.8}{3.8}
& \meanstd{0.3}{0.4}\,/\,\meanstd{0.0}{0.0}
& \meanstd{2.7}{2.9}\,/\,\meanstd{0.0}{0.0}
& \meanstd{12.0}{1.2}
& \meanstd{22.3}{2.1} \\

+PPO
& \meanstd{14.8}{0.8}
& \meanstd{17.5}{1.6}
& \meanstd{0.1}{0.3}\,/\,\meanstd{0.0}{0.0}
& \meanstd{0.3}{0.3}\,/\,\meanstd{0.0}{0.0}
& \meanstd{9.0}{0.4}
& \meanstd{24.9}{0.5} \\

+\methodrl{} (ours)
& \bestmeanstd{24.3}{1.7}
& \bestmeanstd{26.3}{0.5}
& \bestmeanstd{69.0}{2.2}\,/\,\bestmeanstd{52.2}{1.6}
& \bestmeanstd{75.0}{2.0}\,/\,\bestmeanstd{65.5}{3.0}
& \bestmeanstd{20.2}{1.3}
& \bestmeanstd{30.9}{1.6} \\
\bottomrule
\end{tabular}
\end{adjustbox}
\end{table}

\subsection{Does \methodopd Improve On-Policy Distillation?}
\label{sec:exp:opd}

We evaluate whether the step-level advantages produced by our frozen LLM value critic improve on-policy distillation.
As established in \Cref{sec:exp:setup}, the three evaluated methods form a progression in signal granularity.
Pure OPD relies entirely on teacher supervision without environment reward.
Trajectory-gated OPD adds a GRPO policy loss and uses its trajectory-level advantage to gate distillation.
\methodopd finally resolves both the policy-gradient signal and the gate to the step level.

As shown in \cref{tab:main-opd}, \methodopd yields consistent improvements over standard distillation baselines on AppWorld. 
For Qwen3-4B, \methodopd achieves $20.2\%$ success, outperforming both Pure OPD ($17.0\%$) and Trajectory-gated OPD ($19.2\%$). 
A similar trend is observed for Qwen3-8B, where \methodopd reaches $23.3\%$, compared to $22.2\%$ for Pure OPD and $21.2\%$ for Trajectory-gated OPD. 
These results indicate that step-level gating provides a more reliable distillation signal than uniform or trajectory-level approaches.
This generalization advantage is closely aligned with the observed training dynamics. 
As illustrated in \cref{app:fig-opd-curve}, while pure OPD exhibits rapid initial learning, \methodopd surpasses it after approximately $100$ training steps and achieves the highest final training reward across both model scales. 
This pattern suggests that although uniform distillation provides dense early guidance, it eventually restricts the student's policy space. 
By selectively gating supervision, \methodopd avoids this plateau and continues to optimize based on valid student exploration.

Despite these gains, the efficacy of value-guided distillation is limited by the capability of the chosen LLM value critic. 
On BabyAI, all OPD variants, including \methodopd, completely fail to generalize, yielding success rates near $0\%$~(\cref{append:opd-babyai}).
The failure of pure OPD indicates that token-level supervision from Qwen3-235B alone is insufficient for BabyAI.
\methodopd cannot compensate through its value signal, because Qwen3-235B also serves as the frozen LLM critic, and our ablations show that it is a much weaker critic than Claude Opus 4.6 (\cref{sec:ablation}).
We hypothesize that swapping in a more capable LLM critic (e.g., Claude Opus 4.6) for value prediction while keeping the distillation teacher fixed could resolve this, which remains a direction for future work.


\begin{table}[t]
\centering
\captionsetup{justification=raggedright, singlelinecheck=false, skip=0pt}
\begin{minipage}[b]{0.40\textwidth}
\centering
\caption{OPD results on AppWorld. The frozen LLM critic is Qwen3-235B.}
\label{tab:main-opd}
\end{minipage}\hfill
\begin{minipage}[b]{0.28\textwidth}
\centering
\caption{In-context calibration ablation on AppWorld.}
\label{tab:ablation_design_choice}
\end{minipage}\hfill
\begin{minipage}[b]{0.28\textwidth}
\centering
\caption{Choice of frozen LLM critic for \methodrl on AppWorld.}
\label{tab:ablation_critic}
\end{minipage}
\par\vspace{-4pt}
\footnotesize
\setlength{\tabcolsep}{3.5pt}
\begin{minipage}[t]{0.40\textwidth}
\centering
\setlength{\tabcolsep}{2.5pt}
\begin{tabular*}{\linewidth}[t]{@{\extracolsep{\fill}}lcc@{}}
\toprule
Method & Qwen3-4B & Qwen3-8B \\
\midrule
Base & \meanstd{4.2}{0.4} & \meanstd{10.9}{1.0} \\
\midrule
+OPD & \meanstd{17.0}{0.3} & \meanstd{22.2}{0.7} \\
+Trajectory-gated & \meanstd{19.2}{1.0} & \meanstd{21.2}{1.0} \\
+\methodopd{} (ours) & \bestmeanstd{20.2}{1.8} & \bestmeanstd{23.3}{1.6} \\
\bottomrule
\end{tabular*}
\end{minipage}\hfill
\begin{minipage}[t]{0.28\textwidth}
\centering
\begin{tabular*}{\linewidth}[t]{@{\extracolsep{\fill}}lc@{}}
\toprule
Method & Qwen3-8B \\
\midrule
\padrow{}GRPO & \meanstd{19.8}{3.8} \\
\midrule
\padrow{}\methodrl{} (ours) & \bestmeanstd{26.3}{0.5} \\
\padrow{}\quad w/o calibration & \meanstd{22.1}{0.5} \\
\bottomrule
\end{tabular*}
\end{minipage}\hfill
\begin{minipage}[t]{0.28\textwidth}
\centering
\begin{tabular*}{\linewidth}[t]{@{\extracolsep{\fill}}lc@{}}
\toprule
LLM critic & Qwen3-4B \\
\midrule
\padrow{}Claude Opus 4.6 & \bestmeanstd{24.3}{1.7} \\
\midrule
\padrow{}Qwen3-235B & \meanstd{9.8}{2.3} \\
\padrow{}GLM-5.2 & \meanstd{22.3}{4.1} \\
\bottomrule
\end{tabular*}
\end{minipage}
\end{table}

\subsection{Validating Step-Level Value Signals}

To validate the accuracy of the frozen LLM critic, we evaluate whether its predicted step-level value differences align with true changes in the probability of trajectory success. 
Because environments like AppWorld lack step-level, policy-dependent value annotations, we construct reference labels using extensive Monte-Carlo (MC) rollouts.

We use three checkpoints from a single AppWorld training run, taken at approximately $25\%$, $50\%$, and $75\%$ of training to cover the early, middle, and late stages.
For each checkpoint we randomly select $30$ tasks and sample one successful and one failed trajectory per task under that checkpoint, giving $180$ trajectories with $3{,}899$ states in total.
For each recorded state $s_t$, we replay the original action prefix and sample $M=32$ independent continuations from the same checkpoint.
The fraction that succeed is our Monte-Carlo value estimate $V^{\mathrm{MC}}(s_t)$, and $\Delta_t^{\mathrm{MC}}=V^{\mathrm{MC}}(s_{t+1})-V^{\mathrm{MC}}(s_t)$ measures the change across a transition.
We label a transition a \emph{pitfall} when $\Delta_t^{\mathrm{MC}}\leq-3\sigma$ and a \emph{recovery} when $\Delta_t^{\mathrm{MC}}\geq3\sigma$, where $\sigma=0.125$ is the largest possible standard deviation of the difference between two independent estimates at $M=32$ as detailed in~\cref{app:threshold}.
Transitions with $|\Delta_t^{\mathrm{MC}}|<\sigma$ are \emph{MC-neutral}: their estimated change is indistinguishable from sampling noise.
This yields $76$ pitfalls, $38$ recoveries, and $1{,}533$ MC-neutral transitions.

For evaluation, let $\Delta_t^{\mathrm{pred}} = \Vllm(s_{t+1})-\Vllm(s_t)$ denote the predicted value difference.
The group-level normalization (\cref{sec:value-rl}) used during training divides this difference by a positive constant and therefore preserves its sign.
Thus, we consider a step $t$ to be a predicted pitfall when $\Delta_t^{\mathrm{pred}}<0$ and a predicted recovery when $\Delta_t^{\mathrm{pred}}>0$.
Our analysis reveals that the predicted value change has the same sign as the Monte-Carlo change for $72\%$ of pitfalls and $66\%$ of recoveries.
In comparison, it decreases on only $8\%$ of MC-neutral transitions and increases on $22\%$.
These substantial margins provide strong empirical evidence that the frozen LLM critic extracts meaningful, step-level outcome signals rather than random noise.

We further compare \methodrl against a rule-based \emph{error-marker} baseline for pitfall detection. 
This baseline labels a transition as a pitfall simply if the environment output contains text such as \texttt{Traceback}, \texttt{Execution failed}, or \texttt{No code available to execute}. 
The baseline detects only $13\%$ of the MC pitfalls and flags $3\%$ of the MC-neutral transitions. 
This demonstrates that explicit execution failures alone cannot fully account for the pitfall detection achieved by our predicted value signal. 
As illustrated in \cref{app:ex2}, a step may produce an API error yet establish the correct task logic.
The value model correctly identifies such decisions as positive progress, distinguishing the trajectory from other failed rollouts.

\begin{figure}[t]
\centering
\includegraphics[width=\textwidth]{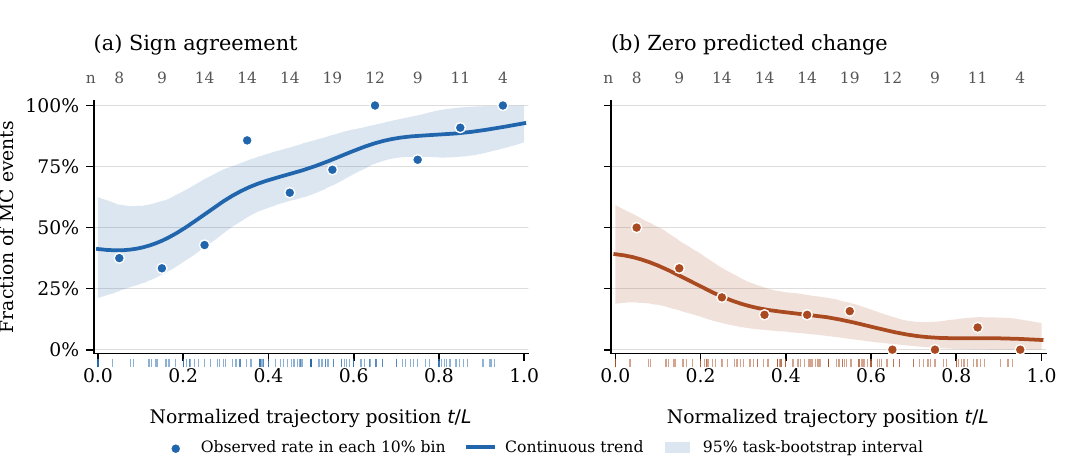}
\caption{
Agreement between predicted value changes and Monte-Carlo events across normalized trajectory position $t/L$, where $L$ is the number of actions.
(a) Fraction of MC events in each position bin for which the predicted value changes in the same direction as the MC estimate: a decrease for pitfalls and an increase for recoveries.
(b) Fraction of MC events for which the model predicts no value change ($\Delta_t^{\mathrm{pred}}=0$), despite an MC increase or decrease large enough to meet the event threshold ($|\Delta_t^{\mathrm{MC}}|\geq3\sigma$).
}
\label{fig:value-agreement}
\end{figure}

To understand where the remaining errors occur, we group the $114$ Monte-Carlo events into ten bins by their position in the trajectory.
As shown in~\cref{fig:value-agreement}, agreement generally increases later in a trajectory.
As a coarse summary, the sign-agreement rate increases from $44\%$ in the first third to $78\%$ in the middle third and $88\%$ in the final third.
In particular, the model assigns no value change to $29\%$ of early events, compared with only $4\%$ of late events, indicating that it often fails to react to events early in a trajectory.
This pattern suggests that the predicted signal becomes more reliable as the trajectory progresses.
We further analyze these early actions where the critic and the Monte-Carlo estimates diverged. 
Early disagreements frequently happen in useful preparatory actions, such as inspecting API documentation, logging in, or retrieving task data. 
In $12$ out of $19$ early disagreements, the LLM critic interprets the action as semantic progress and either maintains or increases its value prediction, whereas the empirical MC value actually decreases. 
This pattern suggests that the frozen LLM critic's value estimates are not strictly on-policy. 
While it correctly identifies useful semantic progress, it overestimates the current student's ability to successfully complete the task.

\subsection{Ablation Studies}
\label{sec:ablation}

\textbf{In-context calibration of the frozen LLM critic.}
Our LLM critic uses information from the rollout group to calibrate its value estimates to the current policy.
The group success rate $\mu_G$ and final reward $R(\tau_i)$ anchor the initial-state and terminal-state values, respectively, while summaries of \(k\) randomly sampled rollouts $\{S(\tau_j)\}_{j\in\mathcal{C}_i}$ provide same-task examples from the current policy.
Because these components jointly ground the critic's expectations in empirical outcomes, we ablate them together by providing only the current trajectory $\tau_i$ to the critic and removing the endpoint anchors.
As shown in \cref{tab:ablation_design_choice}, removing this calibration context reduces the AppWorld success rate from $26.3\%$ to $22.1\%$.

While this ablated variant still outperforms the standard GRPO baseline (\(19.8\%\)), the full method achieves the highest mean success rate, confirming the utility of in-context calibration. 
Although separating the effect of the endpoint anchors from that of the context rollouts, and varying the number of context rollouts, is left to future work, we hypothesize that performance improves with context size. 
Specifically, incorporating more context rollouts should yield a richer, more accurate representation of the current policy, thereby further enhancing the critic's value estimation.

\textbf{Choice of the frozen LLM critic.}
We evaluate how the underlying capability of the frozen LLM critic impacts downstream policy learning.
Using \methodrl, we train Qwen3-4B on AppWorld with different models as the frozen LLM value critic: Qwen3-235B, GLM-5.2~\citep{zeng2026glm}, and Claude Opus 4.6.
All other training settings are fixed, and we train three independent seeds per critic.
As shown in \cref{tab:ablation_critic}, while all evaluated critics successfully improve upon the base model's \(4.2\%\) success rate, the magnitude of this improvement depends strongly on the critic's capability. 
For instance, using Qwen3-235B as the critic yields a modest success rate of \(9.8\%\). 
In contrast, substituting it with more capable models dramatically unlocks the student's learning potential, reaching \(22.3\%\) with GLM-5.2 and \(24.3\%\) with Claude Opus 4.6.
This gap shows that the effectiveness of step-level credit assignment is bottlenecked by the capability of the frozen critic, and it supports our explanation of the \methodopd results on BabyAI in \cref{sec:exp:opd}, where Qwen3-235B serves as the critic.
Furthermore, the comparably high performance of GLM-5.2 and Claude Opus 4.6 suggests that \methodrl generalizes across different frontier model families. 

\section{Conclusion and Discussion}
\label{sec:conclusion}

This work introduces \fullmethodrl (\methodrl), a method for dense credit assignment in LLM agent reinforcement learning. 
Instead of training a separate value model, \methodrl uses a frozen LLM to estimate the value of intermediate states in a completed trajectory. 
It makes these estimates policy-aware by conditioning the LLM on rollout statistics from the current policy via in-context learning. 
By anchoring the initial and terminal values to the group success rate and the actual environment outcome, the consecutive value differences telescope to the group-relative return. 
This process safely redistributes the actual reward across individual steps, providing an advantage signal that distinguishes progress, mistakes, and recoveries without additional environment interactions. 
We also introduce \fullmethodopd (\methodopd), which reuses this step-level advantage to gate on-policy distillation, directing teacher supervision strictly to actions that decrease the estimated value. 
Across agent environments spanning tool use, software engineering, and spatial navigation, \methodrl outperforms GRPO and PPO baselines, while \methodopd improves upon uniform and trajectory-gated distillation baselines.

Despite these strong results, our approach has several limitations. 
First, the method heavily relies on the capability of the frozen LLM critic to judge task progress. 
As shown in our experiments, a weaker critic yields substantially smaller downstream gains. 
For instance, our \methodopd experiments relied on Qwen3-235B as both the teacher and critic. 
Evaluating \methodopd with a stronger critic, such as Claude Opus 4.6, could potentially unlock much higher distillation performance, which we leave to future work.
Second, \methodrl introduces additional inference overhead. 
Querying a large LLM to predict step values for every trajectory during training can be prohibitive when training throughput or compute budget is constrained. 
Future work can investigate reducing this cost, for example, by querying the estimator only when its value signal is highly uncertain or by distilling its step-level knowledge into a smaller, learned critic.
Finally, we designed \methodrl specifically for multi-turn agentic environments, where interactions decompose naturally into textual steps. 
Extending this step-level credit assignment to non-agentic RL domains, such as mathematical reasoning, remains an open and promising direction.

\bibliography{ref}
\bibliographystyle{iclr2027_conference}

\clearpage
\beginappendix
\crefalias{section}{appendix}
\crefalias{subsection}{appendix}
\crefalias{subsubsection}{appendix}

\section{Environment and Task Details}
\label{app:sec1_env}
\subsection{AppWorld}
\label{app:appworld}

\paragraph{Environment.}
AppWorld~\citep{trivedi2024appworld} is a high-fidelity simulator of a personal digital environment in which the agent acts on behalf of a simulated user by writing Python against nine day-to-day applications (\texttt{amazon}, \texttt{file\_system}, \texttt{gmail}, \texttt{phone}, \texttt{simple\_note}, \texttt{splitwise}, \texttt{spotify}, \texttt{todoist}, \texttt{venmo}) exposed through documented APIs.
Two meta-applications complete the interface: \texttt{api\_docs}, which the agent queries to discover apps, list their APIs, and retrieve individual specifications, and \texttt{supervisor}, which holds the user's profile, account credentials, and the \texttt{complete\_task} call that terminates an episode.
The agent must discover the relevant APIs through \texttt{api\_docs}, authenticate by retrieving credentials from \texttt{supervisor}, and only then compose the calls that satisfy the instruction.

\paragraph{Task splits.}
Policy optimization uses only the $90$-task \texttt{train} split ($30$ scenarios $\times$ $3$ variants).
We report final performance on \texttt{test\_challenge} ($417$ tasks), whose scenarios are disjoint from those in \texttt{train}.

\paragraph{Agent interface.}
Each episode is a multi-turn conversation over a stateful Python REPL.
On every turn the policy emits free-form reasoning followed by a single fenced Python code block; the environment executes that block in a session that persists across turns, so variables, imports, and access tokens carry over.
It returns either the captured standard output, an \texttt{Execution successful.} acknowledgement, or the Python traceback on error, with a $30$-second timeout per block.
An episode ends when the agent calls \texttt{apis.supervisor.complete\_task()}, passing \texttt{answer=<value>} for question-answering tasks and no argument for state-changing ones, or when the turn budget is exhausted.
\cref{fig:appworld-traj} shows an example.

\paragraph{Instruction prompt.}
We use the same prompt for both training and evaluation: a static preamble introducing the three \texttt{api\_docs} discovery calls, a $14$-turn demonstration dialogue on one Spotify task, and an eight-item instruction list (e.g., close every code block; reuse variables across turns; inspect an API specification before calling it; write one small code block per turn; page through paginated APIs via \texttt{page\_index}; call \texttt{complete\_task} when done).

\paragraph{Reward.}
Reward is sparse, binary, and awarded only at the end of an episode.
When the agent calls \texttt{complete\_task}, or when the turn budget runs out, AppWorld runs its own validation suite against the resulting database state, and we set $r=1$ if and only if that suite passes in full.
There is no intermediate shaping and no partial credit: the fraction of validation tests passed is logged for analysis but never enters the reward.
Calling \texttt{complete\_task} is not sufficient for success: an episode that declares completion while the validation suite fails receives zero reward, and this is by far the most common failure mode we observe.

\paragraph{Evaluation configuration.}
At evaluation the trained policy is served with vLLM through an OpenAI-compatible endpoint, with the same prompt, action format, and binary success criterion as training.
We sample at temperature $0.6$ and take $3$ rollouts per task, reporting pass@1 over all $417$ \texttt{test\_challenge} tasks.
Two budgets are relaxed relative to training: the turn cap is $70$, and the context cap is $40$k tokens.
Both budgets are held identical across all compared methods.

\begin{figure}[htbp]
\centering
\begin{Verbatim}[fontsize=\scriptsize]
--- Task ---

Using these APIs, now generate code to solve the actual task:
My name is: Kyle Moore. My personal email is kyl_moor@gmail.com
and phone number is 6372953330.
Task: Name the artist most recommended to me on Spotify.

[agent] To find the artist most recommended to me on Spotify, I need to find
        the "recommendations" API. Let me check which APIs Spotify exposes.
        print(apis.api_docs.show_api_descriptions(app_name='spotify'))

[env]   Output: [... "show_recommendations : Get personalized song
        recommendations ..." ...]

[agent] print(apis.api_docs.show_api_doc(app_name='spotify',
                                         api_name='show_recommendations'))

[env]   Output: {"app_name": "spotify", "api_name": "show_recommendations",
        "parameters": [{"name": "access_token", "required": true, ...}]}

[agent] I need to page through all the songs and extract the artist names.
        page_index = 0; songs = []
        while page_index < 10:
            song_page = apis.spotify.show_recommendations(
                access_token=spotify_access_token, page_index=page_index)

[env]   Execution failed. Traceback:
          NameError: name 'spotify_access_token' is not defined

[agent] I need to get the access token first. Let's log in to Spotify.
        spotify_password = [p for p in passwords
                            if p["account_name"] == "spotify"][0]["password"]

[env]   Execution failed. Traceback:
          NameError: name 'passwords' is not defined

[agent] I need to get the passwords first using the `supervisor` app.
        passwords = apis.supervisor.show_account_passwords()

[env]   Execution successful.

                              [ ... ]

[agent] apis.supervisor.complete_task(answer=most_recommended_artist)

[env]   Execution successful.        -> validation suite passes, r = 1
\end{Verbatim}
\caption{
An excerpt from a successful AppWorld episode (Qwen3-4B, task \texttt{287e338\_1}), abridged and reflowed; the enclosing Python fences are omitted for space.
The REPL is stateful, so the agent recovers from the two \texttt{NameError}s by fetching credentials and re-establishing the access token in later turns rather than restarting.
Reward is granted only after \texttt{complete\_task} triggers AppWorld's validation suite.
}
\label{fig:appworld-traj}
\end{figure}

\FloatBarrier
\subsection{BabyAI Task Construction and Evaluation}
\label{app:babyai}
 
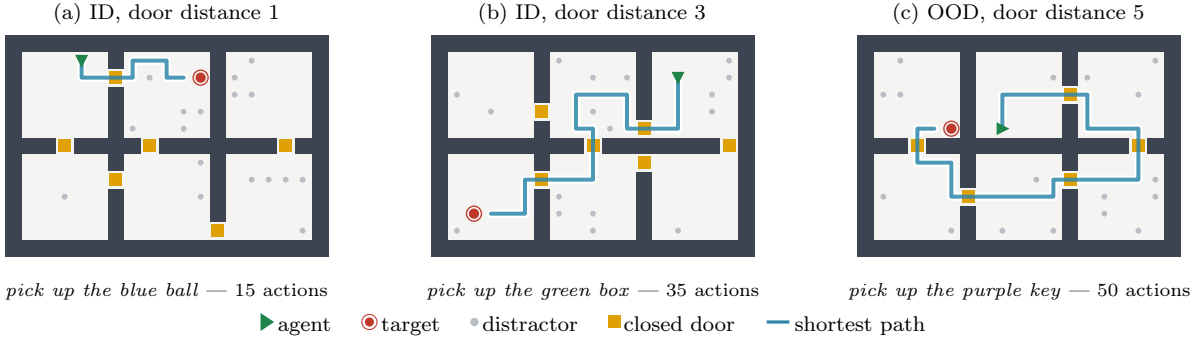
\begin{figure}[t]
\centering
\begin{minipage}[t]{0.32\linewidth}\centering
  {\footnotesize (a) ID, door distance 1}\\[3pt]
  \begin{tikzpicture}[x=\bcell,y=-\bcell,line join=round]
  \fill[bwall] (0,0) rectangle (19,13);
  \fill[bfloor] (1,1) rectangle (2,2);
  \fill[bfloor] (2,1) rectangle (3,2);
  \fill[bfloor] (3,1) rectangle (4,2);
  \fill[bfloor] (4,1) rectangle (5,2);
  \fill[bfloor] (5,1) rectangle (6,2);
  \fill[bfloor] (7,1) rectangle (8,2);
  \fill[bfloor] (8,1) rectangle (9,2);
  \fill[bfloor] (9,1) rectangle (10,2);
  \fill[bfloor] (10,1) rectangle (11,2);
  \fill[bfloor] (11,1) rectangle (12,2);
  \fill[bfloor] (13,1) rectangle (14,2);
  \fill[bfloor] (14,1) rectangle (15,2);
  \fill[bfloor] (15,1) rectangle (16,2);
  \fill[bfloor] (16,1) rectangle (17,2);
  \fill[bfloor] (17,1) rectangle (18,2);
  \fill[bfloor] (1,2) rectangle (2,3);
  \fill[bfloor] (2,2) rectangle (3,3);
  \fill[bfloor] (3,2) rectangle (4,3);
  \fill[bfloor] (4,2) rectangle (5,3);
  \fill[bfloor] (5,2) rectangle (6,3);
  \fill[bfloor] (6,2) rectangle (7,3);
  \fill[bfloor] (7,2) rectangle (8,3);
  \fill[bfloor] (8,2) rectangle (9,3);
  \fill[bfloor] (9,2) rectangle (10,3);
  \fill[bfloor] (10,2) rectangle (11,3);
  \fill[bfloor] (11,2) rectangle (12,3);
  \fill[bfloor] (13,2) rectangle (14,3);
  \fill[bfloor] (14,2) rectangle (15,3);
  \fill[bfloor] (15,2) rectangle (16,3);
  \fill[bfloor] (16,2) rectangle (17,3);
  \fill[bfloor] (17,2) rectangle (18,3);
  \fill[bfloor] (1,3) rectangle (2,4);
  \fill[bfloor] (2,3) rectangle (3,4);
  \fill[bfloor] (3,3) rectangle (4,4);
  \fill[bfloor] (4,3) rectangle (5,4);
  \fill[bfloor] (5,3) rectangle (6,4);
  \fill[bfloor] (7,3) rectangle (8,4);
  \fill[bfloor] (8,3) rectangle (9,4);
  \fill[bfloor] (9,3) rectangle (10,4);
  \fill[bfloor] (10,3) rectangle (11,4);
  \fill[bfloor] (11,3) rectangle (12,4);
  \fill[bfloor] (13,3) rectangle (14,4);
  \fill[bfloor] (14,3) rectangle (15,4);
  \fill[bfloor] (15,3) rectangle (16,4);
  \fill[bfloor] (16,3) rectangle (17,4);
  \fill[bfloor] (17,3) rectangle (18,4);
  \fill[bfloor] (1,4) rectangle (2,5);
  \fill[bfloor] (2,4) rectangle (3,5);
  \fill[bfloor] (3,4) rectangle (4,5);
  \fill[bfloor] (4,4) rectangle (5,5);
  \fill[bfloor] (5,4) rectangle (6,5);
  \fill[bfloor] (7,4) rectangle (8,5);
  \fill[bfloor] (8,4) rectangle (9,5);
  \fill[bfloor] (9,4) rectangle (10,5);
  \fill[bfloor] (10,4) rectangle (11,5);
  \fill[bfloor] (11,4) rectangle (12,5);
  \fill[bfloor] (13,4) rectangle (14,5);
  \fill[bfloor] (14,4) rectangle (15,5);
  \fill[bfloor] (15,4) rectangle (16,5);
  \fill[bfloor] (16,4) rectangle (17,5);
  \fill[bfloor] (17,4) rectangle (18,5);
  \fill[bfloor] (1,5) rectangle (2,6);
  \fill[bfloor] (2,5) rectangle (3,6);
  \fill[bfloor] (3,5) rectangle (4,6);
  \fill[bfloor] (4,5) rectangle (5,6);
  \fill[bfloor] (5,5) rectangle (6,6);
  \fill[bfloor] (7,5) rectangle (8,6);
  \fill[bfloor] (8,5) rectangle (9,6);
  \fill[bfloor] (9,5) rectangle (10,6);
  \fill[bfloor] (10,5) rectangle (11,6);
  \fill[bfloor] (11,5) rectangle (12,6);
  \fill[bfloor] (13,5) rectangle (14,6);
  \fill[bfloor] (14,5) rectangle (15,6);
  \fill[bfloor] (15,5) rectangle (16,6);
  \fill[bfloor] (16,5) rectangle (17,6);
  \fill[bfloor] (17,5) rectangle (18,6);
  \fill[bfloor] (3,6) rectangle (4,7);
  \fill[bfloor] (8,6) rectangle (9,7);
  \fill[bfloor] (16,6) rectangle (17,7);
  \fill[bfloor] (1,7) rectangle (2,8);
  \fill[bfloor] (2,7) rectangle (3,8);
  \fill[bfloor] (3,7) rectangle (4,8);
  \fill[bfloor] (4,7) rectangle (5,8);
  \fill[bfloor] (5,7) rectangle (6,8);
  \fill[bfloor] (7,7) rectangle (8,8);
  \fill[bfloor] (8,7) rectangle (9,8);
  \fill[bfloor] (9,7) rectangle (10,8);
  \fill[bfloor] (10,7) rectangle (11,8);
  \fill[bfloor] (11,7) rectangle (12,8);
  \fill[bfloor] (13,7) rectangle (14,8);
  \fill[bfloor] (14,7) rectangle (15,8);
  \fill[bfloor] (15,7) rectangle (16,8);
  \fill[bfloor] (16,7) rectangle (17,8);
  \fill[bfloor] (17,7) rectangle (18,8);
  \fill[bfloor] (1,8) rectangle (2,9);
  \fill[bfloor] (2,8) rectangle (3,9);
  \fill[bfloor] (3,8) rectangle (4,9);
  \fill[bfloor] (4,8) rectangle (5,9);
  \fill[bfloor] (5,8) rectangle (6,9);
  \fill[bfloor] (6,8) rectangle (7,9);
  \fill[bfloor] (7,8) rectangle (8,9);
  \fill[bfloor] (8,8) rectangle (9,9);
  \fill[bfloor] (9,8) rectangle (10,9);
  \fill[bfloor] (10,8) rectangle (11,9);
  \fill[bfloor] (11,8) rectangle (12,9);
  \fill[bfloor] (13,8) rectangle (14,9);
  \fill[bfloor] (14,8) rectangle (15,9);
  \fill[bfloor] (15,8) rectangle (16,9);
  \fill[bfloor] (16,8) rectangle (17,9);
  \fill[bfloor] (17,8) rectangle (18,9);
  \fill[bfloor] (1,9) rectangle (2,10);
  \fill[bfloor] (2,9) rectangle (3,10);
  \fill[bfloor] (3,9) rectangle (4,10);
  \fill[bfloor] (4,9) rectangle (5,10);
  \fill[bfloor] (5,9) rectangle (6,10);
  \fill[bfloor] (7,9) rectangle (8,10);
  \fill[bfloor] (8,9) rectangle (9,10);
  \fill[bfloor] (9,9) rectangle (10,10);
  \fill[bfloor] (10,9) rectangle (11,10);
  \fill[bfloor] (11,9) rectangle (12,10);
  \fill[bfloor] (13,9) rectangle (14,10);
  \fill[bfloor] (14,9) rectangle (15,10);
  \fill[bfloor] (15,9) rectangle (16,10);
  \fill[bfloor] (16,9) rectangle (17,10);
  \fill[bfloor] (17,9) rectangle (18,10);
  \fill[bfloor] (1,10) rectangle (2,11);
  \fill[bfloor] (2,10) rectangle (3,11);
  \fill[bfloor] (3,10) rectangle (4,11);
  \fill[bfloor] (4,10) rectangle (5,11);
  \fill[bfloor] (5,10) rectangle (6,11);
  \fill[bfloor] (7,10) rectangle (8,11);
  \fill[bfloor] (8,10) rectangle (9,11);
  \fill[bfloor] (9,10) rectangle (10,11);
  \fill[bfloor] (10,10) rectangle (11,11);
  \fill[bfloor] (11,10) rectangle (12,11);
  \fill[bfloor] (13,10) rectangle (14,11);
  \fill[bfloor] (14,10) rectangle (15,11);
  \fill[bfloor] (15,10) rectangle (16,11);
  \fill[bfloor] (16,10) rectangle (17,11);
  \fill[bfloor] (17,10) rectangle (18,11);
  \fill[bfloor] (1,11) rectangle (2,12);
  \fill[bfloor] (2,11) rectangle (3,12);
  \fill[bfloor] (3,11) rectangle (4,12);
  \fill[bfloor] (4,11) rectangle (5,12);
  \fill[bfloor] (5,11) rectangle (6,12);
  \fill[bfloor] (7,11) rectangle (8,12);
  \fill[bfloor] (8,11) rectangle (9,12);
  \fill[bfloor] (9,11) rectangle (10,12);
  \fill[bfloor] (10,11) rectangle (11,12);
  \fill[bfloor] (11,11) rectangle (12,12);
  \fill[bfloor] (12,11) rectangle (13,12);
  \fill[bfloor] (13,11) rectangle (14,12);
  \fill[bfloor] (14,11) rectangle (15,12);
  \fill[bfloor] (15,11) rectangle (16,12);
  \fill[bfloor] (16,11) rectangle (17,12);
  \fill[bfloor] (17,11) rectangle (18,12);
  \draw[bpathcase] (4.5,1.5) -- (4.5,2.5) -- (5.5,2.5) -- (6.5,2.5) -- (7.5,2.5) -- (7.5,1.5) -- (8.5,1.5) -- (9.5,1.5) -- (9.5,2.5) -- (10.5,2.5);
  \fill[bdoor] (6.12,2.12) rectangle (6.88,2.88);
  \fill[bdoor] (3.12,6.12) rectangle (3.88,6.88);
  \fill[bdoor] (8.12,6.12) rectangle (8.88,6.88);
  \fill[bdoor] (16.12,6.12) rectangle (16.88,6.88);
  \fill[bdoor] (6.12,8.12) rectangle (6.88,8.88);
  \fill[bdoor] (12.12,11.12) rectangle (12.88,11.88);
  \draw[bpath] (4.5,1.5) -- (4.5,2.5) -- (5.5,2.5) -- (6.5,2.5) -- (7.5,2.5) -- (7.5,1.5) -- (8.5,1.5) -- (9.5,1.5) -- (9.5,2.5) -- (10.5,2.5);
  \fill[bobj] (14.5,1.5) circle (0.17);
  \fill[bobj] (8.5,2.5) circle (0.17);
  \fill[bobj] (13.5,2.5) circle (0.17);
  \fill[bobj] (13.5,3.5) circle (0.17);
  \fill[bobj] (14.5,3.5) circle (0.17);
  \fill[bobj] (10.5,4.5) circle (0.17);
  \fill[bobj] (11.5,4.5) circle (0.17);
  \fill[bobj] (7.5,5.5) circle (0.17);
  \fill[bobj] (10.5,5.5) circle (0.17);
  \fill[bobj] (11.5,7.5) circle (0.17);
  \fill[bobj] (14.5,8.5) circle (0.17);
  \fill[bobj] (15.5,8.5) circle (0.17);
  \fill[bobj] (16.5,8.5) circle (0.17);
  \fill[bobj] (17.5,8.5) circle (0.17);
  \fill[bobj] (3.5,9.5) circle (0.17);
  \fill[bobj] (11.5,9.5) circle (0.17);
  \fill[bobj] (17.5,11.5) circle (0.17);
  \fill[btarget] (11.5,2.5) circle (0.30);
  \draw[btargetring] (11.5,2.5) circle (0.44);
  \fill[bagent,rotate around={-90:(4.5,1.5)}] (4.92,1.50) -- (4.16,1.12) -- (4.16,1.88) -- cycle;
\end{tikzpicture}\\[3pt]
  {\scriptsize \emph{pick up the blue ball} --- 15 actions}
\end{minipage}\hfill
\begin{minipage}[t]{0.32\linewidth}\centering
  {\footnotesize (b) ID, door distance 3}\\[3pt]
  \begin{tikzpicture}[x=\bcell,y=-\bcell,line join=round]
  \fill[bwall] (0,0) rectangle (19,13);
  \fill[bfloor] (1,1) rectangle (2,2);
  \fill[bfloor] (2,1) rectangle (3,2);
  \fill[bfloor] (3,1) rectangle (4,2);
  \fill[bfloor] (4,1) rectangle (5,2);
  \fill[bfloor] (5,1) rectangle (6,2);
  \fill[bfloor] (7,1) rectangle (8,2);
  \fill[bfloor] (8,1) rectangle (9,2);
  \fill[bfloor] (9,1) rectangle (10,2);
  \fill[bfloor] (10,1) rectangle (11,2);
  \fill[bfloor] (11,1) rectangle (12,2);
  \fill[bfloor] (13,1) rectangle (14,2);
  \fill[bfloor] (14,1) rectangle (15,2);
  \fill[bfloor] (15,1) rectangle (16,2);
  \fill[bfloor] (16,1) rectangle (17,2);
  \fill[bfloor] (17,1) rectangle (18,2);
  \fill[bfloor] (1,2) rectangle (2,3);
  \fill[bfloor] (2,2) rectangle (3,3);
  \fill[bfloor] (3,2) rectangle (4,3);
  \fill[bfloor] (4,2) rectangle (5,3);
  \fill[bfloor] (5,2) rectangle (6,3);
  \fill[bfloor] (7,2) rectangle (8,3);
  \fill[bfloor] (8,2) rectangle (9,3);
  \fill[bfloor] (9,2) rectangle (10,3);
  \fill[bfloor] (10,2) rectangle (11,3);
  \fill[bfloor] (11,2) rectangle (12,3);
  \fill[bfloor] (13,2) rectangle (14,3);
  \fill[bfloor] (14,2) rectangle (15,3);
  \fill[bfloor] (15,2) rectangle (16,3);
  \fill[bfloor] (16,2) rectangle (17,3);
  \fill[bfloor] (17,2) rectangle (18,3);
  \fill[bfloor] (1,3) rectangle (2,4);
  \fill[bfloor] (2,3) rectangle (3,4);
  \fill[bfloor] (3,3) rectangle (4,4);
  \fill[bfloor] (4,3) rectangle (5,4);
  \fill[bfloor] (5,3) rectangle (6,4);
  \fill[bfloor] (7,3) rectangle (8,4);
  \fill[bfloor] (8,3) rectangle (9,4);
  \fill[bfloor] (9,3) rectangle (10,4);
  \fill[bfloor] (10,3) rectangle (11,4);
  \fill[bfloor] (11,3) rectangle (12,4);
  \fill[bfloor] (13,3) rectangle (14,4);
  \fill[bfloor] (14,3) rectangle (15,4);
  \fill[bfloor] (15,3) rectangle (16,4);
  \fill[bfloor] (16,3) rectangle (17,4);
  \fill[bfloor] (17,3) rectangle (18,4);
  \fill[bfloor] (1,4) rectangle (2,5);
  \fill[bfloor] (2,4) rectangle (3,5);
  \fill[bfloor] (3,4) rectangle (4,5);
  \fill[bfloor] (4,4) rectangle (5,5);
  \fill[bfloor] (5,4) rectangle (6,5);
  \fill[bfloor] (6,4) rectangle (7,5);
  \fill[bfloor] (7,4) rectangle (8,5);
  \fill[bfloor] (8,4) rectangle (9,5);
  \fill[bfloor] (9,4) rectangle (10,5);
  \fill[bfloor] (10,4) rectangle (11,5);
  \fill[bfloor] (11,4) rectangle (12,5);
  \fill[bfloor] (13,4) rectangle (14,5);
  \fill[bfloor] (14,4) rectangle (15,5);
  \fill[bfloor] (15,4) rectangle (16,5);
  \fill[bfloor] (16,4) rectangle (17,5);
  \fill[bfloor] (17,4) rectangle (18,5);
  \fill[bfloor] (1,5) rectangle (2,6);
  \fill[bfloor] (2,5) rectangle (3,6);
  \fill[bfloor] (3,5) rectangle (4,6);
  \fill[bfloor] (4,5) rectangle (5,6);
  \fill[bfloor] (5,5) rectangle (6,6);
  \fill[bfloor] (7,5) rectangle (8,6);
  \fill[bfloor] (8,5) rectangle (9,6);
  \fill[bfloor] (9,5) rectangle (10,6);
  \fill[bfloor] (10,5) rectangle (11,6);
  \fill[bfloor] (11,5) rectangle (12,6);
  \fill[bfloor] (12,5) rectangle (13,6);
  \fill[bfloor] (13,5) rectangle (14,6);
  \fill[bfloor] (14,5) rectangle (15,6);
  \fill[bfloor] (15,5) rectangle (16,6);
  \fill[bfloor] (16,5) rectangle (17,6);
  \fill[bfloor] (17,5) rectangle (18,6);
  \fill[bfloor] (9,6) rectangle (10,7);
  \fill[bfloor] (17,6) rectangle (18,7);
  \fill[bfloor] (1,7) rectangle (2,8);
  \fill[bfloor] (2,7) rectangle (3,8);
  \fill[bfloor] (3,7) rectangle (4,8);
  \fill[bfloor] (4,7) rectangle (5,8);
  \fill[bfloor] (5,7) rectangle (6,8);
  \fill[bfloor] (7,7) rectangle (8,8);
  \fill[bfloor] (8,7) rectangle (9,8);
  \fill[bfloor] (9,7) rectangle (10,8);
  \fill[bfloor] (10,7) rectangle (11,8);
  \fill[bfloor] (11,7) rectangle (12,8);
  \fill[bfloor] (12,7) rectangle (13,8);
  \fill[bfloor] (13,7) rectangle (14,8);
  \fill[bfloor] (14,7) rectangle (15,8);
  \fill[bfloor] (15,7) rectangle (16,8);
  \fill[bfloor] (16,7) rectangle (17,8);
  \fill[bfloor] (17,7) rectangle (18,8);
  \fill[bfloor] (1,8) rectangle (2,9);
  \fill[bfloor] (2,8) rectangle (3,9);
  \fill[bfloor] (3,8) rectangle (4,9);
  \fill[bfloor] (4,8) rectangle (5,9);
  \fill[bfloor] (5,8) rectangle (6,9);
  \fill[bfloor] (6,8) rectangle (7,9);
  \fill[bfloor] (7,8) rectangle (8,9);
  \fill[bfloor] (8,8) rectangle (9,9);
  \fill[bfloor] (9,8) rectangle (10,9);
  \fill[bfloor] (10,8) rectangle (11,9);
  \fill[bfloor] (11,8) rectangle (12,9);
  \fill[bfloor] (13,8) rectangle (14,9);
  \fill[bfloor] (14,8) rectangle (15,9);
  \fill[bfloor] (15,8) rectangle (16,9);
  \fill[bfloor] (16,8) rectangle (17,9);
  \fill[bfloor] (17,8) rectangle (18,9);
  \fill[bfloor] (1,9) rectangle (2,10);
  \fill[bfloor] (2,9) rectangle (3,10);
  \fill[bfloor] (3,9) rectangle (4,10);
  \fill[bfloor] (4,9) rectangle (5,10);
  \fill[bfloor] (5,9) rectangle (6,10);
  \fill[bfloor] (7,9) rectangle (8,10);
  \fill[bfloor] (8,9) rectangle (9,10);
  \fill[bfloor] (9,9) rectangle (10,10);
  \fill[bfloor] (10,9) rectangle (11,10);
  \fill[bfloor] (11,9) rectangle (12,10);
  \fill[bfloor] (13,9) rectangle (14,10);
  \fill[bfloor] (14,9) rectangle (15,10);
  \fill[bfloor] (15,9) rectangle (16,10);
  \fill[bfloor] (16,9) rectangle (17,10);
  \fill[bfloor] (17,9) rectangle (18,10);
  \fill[bfloor] (1,10) rectangle (2,11);
  \fill[bfloor] (2,10) rectangle (3,11);
  \fill[bfloor] (3,10) rectangle (4,11);
  \fill[bfloor] (4,10) rectangle (5,11);
  \fill[bfloor] (5,10) rectangle (6,11);
  \fill[bfloor] (7,10) rectangle (8,11);
  \fill[bfloor] (8,10) rectangle (9,11);
  \fill[bfloor] (9,10) rectangle (10,11);
  \fill[bfloor] (10,10) rectangle (11,11);
  \fill[bfloor] (11,10) rectangle (12,11);
  \fill[bfloor] (13,10) rectangle (14,11);
  \fill[bfloor] (14,10) rectangle (15,11);
  \fill[bfloor] (15,10) rectangle (16,11);
  \fill[bfloor] (16,10) rectangle (17,11);
  \fill[bfloor] (17,10) rectangle (18,11);
  \fill[bfloor] (1,11) rectangle (2,12);
  \fill[bfloor] (2,11) rectangle (3,12);
  \fill[bfloor] (3,11) rectangle (4,12);
  \fill[bfloor] (4,11) rectangle (5,12);
  \fill[bfloor] (5,11) rectangle (6,12);
  \fill[bfloor] (7,11) rectangle (8,12);
  \fill[bfloor] (8,11) rectangle (9,12);
  \fill[bfloor] (9,11) rectangle (10,12);
  \fill[bfloor] (10,11) rectangle (11,12);
  \fill[bfloor] (11,11) rectangle (12,12);
  \fill[bfloor] (13,11) rectangle (14,12);
  \fill[bfloor] (14,11) rectangle (15,12);
  \fill[bfloor] (15,11) rectangle (16,12);
  \fill[bfloor] (16,11) rectangle (17,12);
  \fill[bfloor] (17,11) rectangle (18,12);
  \draw[bpathcase] (14.5,2.5) -- (14.5,3.5) -- (14.5,4.5) -- (14.5,5.5) -- (13.5,5.5) -- (12.5,5.5) -- (11.5,5.5) -- (11.5,4.5) -- (11.5,3.5) -- (10.5,3.5) -- (9.5,3.5) -- (8.5,3.5) -- (8.5,4.5) -- (8.5,5.5) -- (9.5,5.5) -- (9.5,6.5) -- (9.5,7.5) -- (9.5,8.5) -- (8.5,8.5) -- (7.5,8.5) -- (6.5,8.5) -- (5.5,8.5) -- (5.5,9.5) -- (5.5,10.5) -- (4.5,10.5) -- (3.5,10.5);
  \fill[bdoor] (6.12,4.12) rectangle (6.88,4.88);
  \fill[bdoor] (12.12,5.12) rectangle (12.88,5.88);
  \fill[bdoor] (9.12,6.12) rectangle (9.88,6.88);
  \fill[bdoor] (17.12,6.12) rectangle (17.88,6.88);
  \fill[bdoor] (12.12,7.12) rectangle (12.88,7.88);
  \fill[bdoor] (6.12,8.12) rectangle (6.88,8.88);
  \draw[bpath] (14.5,2.5) -- (14.5,3.5) -- (14.5,4.5) -- (14.5,5.5) -- (13.5,5.5) -- (12.5,5.5) -- (11.5,5.5) -- (11.5,4.5) -- (11.5,3.5) -- (10.5,3.5) -- (9.5,3.5) -- (8.5,3.5) -- (8.5,4.5) -- (8.5,5.5) -- (9.5,5.5) -- (9.5,6.5) -- (9.5,7.5) -- (9.5,8.5) -- (8.5,8.5) -- (7.5,8.5) -- (6.5,8.5) -- (5.5,8.5) -- (5.5,9.5) -- (5.5,10.5) -- (4.5,10.5) -- (3.5,10.5);
  \fill[bobj] (7.5,1.5) circle (0.17);
  \fill[bobj] (17.5,1.5) circle (0.17);
  \fill[bobj] (9.5,2.5) circle (0.17);
  \fill[bobj] (11.5,2.5) circle (0.17);
  \fill[bobj] (17.5,2.5) circle (0.17);
  \fill[bobj] (1.5,3.5) circle (0.17);
  \fill[bobj] (16.5,3.5) circle (0.17);
  \fill[bobj] (3.5,4.5) circle (0.17);
  \fill[bobj] (9.5,4.5) circle (0.17);
  \fill[bobj] (16.5,4.5) circle (0.17);
  \fill[bobj] (10.5,5.5) circle (0.17);
  \fill[bobj] (7.5,9.5) circle (0.17);
  \fill[bobj] (7.5,10.5) circle (0.17);
  \fill[bobj] (9.5,10.5) circle (0.17);
  \fill[bobj] (1.5,11.5) circle (0.17);
  \fill[bobj] (9.5,11.5) circle (0.17);
  \fill[bobj] (14.5,11.5) circle (0.17);
  \fill[btarget] (2.5,10.5) circle (0.30);
  \draw[btargetring] (2.5,10.5) circle (0.44);
  \fill[bagent,rotate around={-90:(14.5,2.5)}] (14.92,2.50) -- (14.16,2.12) -- (14.16,2.88) -- cycle;
\end{tikzpicture}\\[3pt]
  {\scriptsize \emph{pick up the green box} --- 35 actions}
\end{minipage}\hfill
\begin{minipage}[t]{0.32\linewidth}\centering
  {\footnotesize (c) OOD, door distance 5}\\[3pt]
  \begin{tikzpicture}[x=\bcell,y=-\bcell,line join=round]
  \fill[bwall] (0,0) rectangle (19,13);
  \fill[bfloor] (1,1) rectangle (2,2);
  \fill[bfloor] (2,1) rectangle (3,2);
  \fill[bfloor] (3,1) rectangle (4,2);
  \fill[bfloor] (4,1) rectangle (5,2);
  \fill[bfloor] (5,1) rectangle (6,2);
  \fill[bfloor] (7,1) rectangle (8,2);
  \fill[bfloor] (8,1) rectangle (9,2);
  \fill[bfloor] (9,1) rectangle (10,2);
  \fill[bfloor] (10,1) rectangle (11,2);
  \fill[bfloor] (11,1) rectangle (12,2);
  \fill[bfloor] (13,1) rectangle (14,2);
  \fill[bfloor] (14,1) rectangle (15,2);
  \fill[bfloor] (15,1) rectangle (16,2);
  \fill[bfloor] (16,1) rectangle (17,2);
  \fill[bfloor] (17,1) rectangle (18,2);
  \fill[bfloor] (1,2) rectangle (2,3);
  \fill[bfloor] (2,2) rectangle (3,3);
  \fill[bfloor] (3,2) rectangle (4,3);
  \fill[bfloor] (4,2) rectangle (5,3);
  \fill[bfloor] (5,2) rectangle (6,3);
  \fill[bfloor] (7,2) rectangle (8,3);
  \fill[bfloor] (8,2) rectangle (9,3);
  \fill[bfloor] (9,2) rectangle (10,3);
  \fill[bfloor] (10,2) rectangle (11,3);
  \fill[bfloor] (11,2) rectangle (12,3);
  \fill[bfloor] (13,2) rectangle (14,3);
  \fill[bfloor] (14,2) rectangle (15,3);
  \fill[bfloor] (15,2) rectangle (16,3);
  \fill[bfloor] (16,2) rectangle (17,3);
  \fill[bfloor] (17,2) rectangle (18,3);
  \fill[bfloor] (1,3) rectangle (2,4);
  \fill[bfloor] (2,3) rectangle (3,4);
  \fill[bfloor] (3,3) rectangle (4,4);
  \fill[bfloor] (4,3) rectangle (5,4);
  \fill[bfloor] (5,3) rectangle (6,4);
  \fill[bfloor] (7,3) rectangle (8,4);
  \fill[bfloor] (8,3) rectangle (9,4);
  \fill[bfloor] (9,3) rectangle (10,4);
  \fill[bfloor] (10,3) rectangle (11,4);
  \fill[bfloor] (11,3) rectangle (12,4);
  \fill[bfloor] (12,3) rectangle (13,4);
  \fill[bfloor] (13,3) rectangle (14,4);
  \fill[bfloor] (14,3) rectangle (15,4);
  \fill[bfloor] (15,3) rectangle (16,4);
  \fill[bfloor] (16,3) rectangle (17,4);
  \fill[bfloor] (17,3) rectangle (18,4);
  \fill[bfloor] (1,4) rectangle (2,5);
  \fill[bfloor] (2,4) rectangle (3,5);
  \fill[bfloor] (3,4) rectangle (4,5);
  \fill[bfloor] (4,4) rectangle (5,5);
  \fill[bfloor] (5,4) rectangle (6,5);
  \fill[bfloor] (7,4) rectangle (8,5);
  \fill[bfloor] (8,4) rectangle (9,5);
  \fill[bfloor] (9,4) rectangle (10,5);
  \fill[bfloor] (10,4) rectangle (11,5);
  \fill[bfloor] (11,4) rectangle (12,5);
  \fill[bfloor] (13,4) rectangle (14,5);
  \fill[bfloor] (14,4) rectangle (15,5);
  \fill[bfloor] (15,4) rectangle (16,5);
  \fill[bfloor] (16,4) rectangle (17,5);
  \fill[bfloor] (17,4) rectangle (18,5);
  \fill[bfloor] (1,5) rectangle (2,6);
  \fill[bfloor] (2,5) rectangle (3,6);
  \fill[bfloor] (3,5) rectangle (4,6);
  \fill[bfloor] (4,5) rectangle (5,6);
  \fill[bfloor] (5,5) rectangle (6,6);
  \fill[bfloor] (7,5) rectangle (8,6);
  \fill[bfloor] (8,5) rectangle (9,6);
  \fill[bfloor] (9,5) rectangle (10,6);
  \fill[bfloor] (10,5) rectangle (11,6);
  \fill[bfloor] (11,5) rectangle (12,6);
  \fill[bfloor] (13,5) rectangle (14,6);
  \fill[bfloor] (14,5) rectangle (15,6);
  \fill[bfloor] (15,5) rectangle (16,6);
  \fill[bfloor] (16,5) rectangle (17,6);
  \fill[bfloor] (17,5) rectangle (18,6);
  \fill[bfloor] (3,6) rectangle (4,7);
  \fill[bfloor] (16,6) rectangle (17,7);
  \fill[bfloor] (1,7) rectangle (2,8);
  \fill[bfloor] (2,7) rectangle (3,8);
  \fill[bfloor] (3,7) rectangle (4,8);
  \fill[bfloor] (4,7) rectangle (5,8);
  \fill[bfloor] (5,7) rectangle (6,8);
  \fill[bfloor] (7,7) rectangle (8,8);
  \fill[bfloor] (8,7) rectangle (9,8);
  \fill[bfloor] (9,7) rectangle (10,8);
  \fill[bfloor] (10,7) rectangle (11,8);
  \fill[bfloor] (11,7) rectangle (12,8);
  \fill[bfloor] (13,7) rectangle (14,8);
  \fill[bfloor] (14,7) rectangle (15,8);
  \fill[bfloor] (15,7) rectangle (16,8);
  \fill[bfloor] (16,7) rectangle (17,8);
  \fill[bfloor] (17,7) rectangle (18,8);
  \fill[bfloor] (1,8) rectangle (2,9);
  \fill[bfloor] (2,8) rectangle (3,9);
  \fill[bfloor] (3,8) rectangle (4,9);
  \fill[bfloor] (4,8) rectangle (5,9);
  \fill[bfloor] (5,8) rectangle (6,9);
  \fill[bfloor] (7,8) rectangle (8,9);
  \fill[bfloor] (8,8) rectangle (9,9);
  \fill[bfloor] (9,8) rectangle (10,9);
  \fill[bfloor] (10,8) rectangle (11,9);
  \fill[bfloor] (11,8) rectangle (12,9);
  \fill[bfloor] (12,8) rectangle (13,9);
  \fill[bfloor] (13,8) rectangle (14,9);
  \fill[bfloor] (14,8) rectangle (15,9);
  \fill[bfloor] (15,8) rectangle (16,9);
  \fill[bfloor] (16,8) rectangle (17,9);
  \fill[bfloor] (17,8) rectangle (18,9);
  \fill[bfloor] (1,9) rectangle (2,10);
  \fill[bfloor] (2,9) rectangle (3,10);
  \fill[bfloor] (3,9) rectangle (4,10);
  \fill[bfloor] (4,9) rectangle (5,10);
  \fill[bfloor] (5,9) rectangle (6,10);
  \fill[bfloor] (6,9) rectangle (7,10);
  \fill[bfloor] (7,9) rectangle (8,10);
  \fill[bfloor] (8,9) rectangle (9,10);
  \fill[bfloor] (9,9) rectangle (10,10);
  \fill[bfloor] (10,9) rectangle (11,10);
  \fill[bfloor] (11,9) rectangle (12,10);
  \fill[bfloor] (13,9) rectangle (14,10);
  \fill[bfloor] (14,9) rectangle (15,10);
  \fill[bfloor] (15,9) rectangle (16,10);
  \fill[bfloor] (16,9) rectangle (17,10);
  \fill[bfloor] (17,9) rectangle (18,10);
  \fill[bfloor] (1,10) rectangle (2,11);
  \fill[bfloor] (2,10) rectangle (3,11);
  \fill[bfloor] (3,10) rectangle (4,11);
  \fill[bfloor] (4,10) rectangle (5,11);
  \fill[bfloor] (5,10) rectangle (6,11);
  \fill[bfloor] (7,10) rectangle (8,11);
  \fill[bfloor] (8,10) rectangle (9,11);
  \fill[bfloor] (9,10) rectangle (10,11);
  \fill[bfloor] (10,10) rectangle (11,11);
  \fill[bfloor] (11,10) rectangle (12,11);
  \fill[bfloor] (13,10) rectangle (14,11);
  \fill[bfloor] (14,10) rectangle (15,11);
  \fill[bfloor] (15,10) rectangle (16,11);
  \fill[bfloor] (16,10) rectangle (17,11);
  \fill[bfloor] (17,10) rectangle (18,11);
  \fill[bfloor] (1,11) rectangle (2,12);
  \fill[bfloor] (2,11) rectangle (3,12);
  \fill[bfloor] (3,11) rectangle (4,12);
  \fill[bfloor] (4,11) rectangle (5,12);
  \fill[bfloor] (5,11) rectangle (6,12);
  \fill[bfloor] (7,11) rectangle (8,12);
  \fill[bfloor] (8,11) rectangle (9,12);
  \fill[bfloor] (9,11) rectangle (10,12);
  \fill[bfloor] (10,11) rectangle (11,12);
  \fill[bfloor] (11,11) rectangle (12,12);
  \fill[bfloor] (13,11) rectangle (14,12);
  \fill[bfloor] (14,11) rectangle (15,12);
  \fill[bfloor] (15,11) rectangle (16,12);
  \fill[bfloor] (16,11) rectangle (17,12);
  \fill[bfloor] (17,11) rectangle (18,12);
  \draw[bpathcase] (8.5,5.5) -- (8.5,4.5) -- (8.5,3.5) -- (9.5,3.5) -- (10.5,3.5) -- (11.5,3.5) -- (12.5,3.5) -- (13.5,3.5) -- (13.5,4.5) -- (13.5,5.5) -- (14.5,5.5) -- (15.5,5.5) -- (16.5,5.5) -- (16.5,6.5) -- (16.5,7.5) -- (16.5,8.5) -- (15.5,8.5) -- (14.5,8.5) -- (13.5,8.5) -- (12.5,8.5) -- (11.5,8.5) -- (11.5,9.5) -- (10.5,9.5) -- (9.5,9.5) -- (8.5,9.5) -- (7.5,9.5) -- (6.5,9.5) -- (5.5,9.5) -- (5.5,8.5) -- (5.5,7.5) -- (4.5,7.5) -- (3.5,7.5) -- (3.5,6.5) -- (3.5,5.5) -- (4.5,5.5);
  \fill[bdoor] (12.12,3.12) rectangle (12.88,3.88);
  \fill[bdoor] (3.12,6.12) rectangle (3.88,6.88);
  \fill[bdoor] (16.12,6.12) rectangle (16.88,6.88);
  \fill[bdoor] (12.12,8.12) rectangle (12.88,8.88);
  \fill[bdoor] (6.12,9.12) rectangle (6.88,9.88);
  \draw[bpath] (8.5,5.5) -- (8.5,4.5) -- (8.5,3.5) -- (9.5,3.5) -- (10.5,3.5) -- (11.5,3.5) -- (12.5,3.5) -- (13.5,3.5) -- (13.5,4.5) -- (13.5,5.5) -- (14.5,5.5) -- (15.5,5.5) -- (16.5,5.5) -- (16.5,6.5) -- (16.5,7.5) -- (16.5,8.5) -- (15.5,8.5) -- (14.5,8.5) -- (13.5,8.5) -- (12.5,8.5) -- (11.5,8.5) -- (11.5,9.5) -- (10.5,9.5) -- (9.5,9.5) -- (8.5,9.5) -- (7.5,9.5) -- (6.5,9.5) -- (5.5,9.5) -- (5.5,8.5) -- (5.5,7.5) -- (4.5,7.5) -- (3.5,7.5) -- (3.5,6.5) -- (3.5,5.5) -- (4.5,5.5);
  \fill[bobj] (2.5,1.5) circle (0.17);
  \fill[bobj] (17.5,1.5) circle (0.17);
  \fill[bobj] (11.5,2.5) circle (0.17);
  \fill[bobj] (13.5,2.5) circle (0.17);
  \fill[bobj] (1.5,3.5) circle (0.17);
  \fill[bobj] (2.5,3.5) circle (0.17);
  \fill[bobj] (16.5,3.5) circle (0.17);
  \fill[bobj] (10.5,8.5) circle (0.17);
  \fill[bobj] (17.5,8.5) circle (0.17);
  \fill[bobj] (1.5,9.5) circle (0.17);
  \fill[bobj] (14.5,9.5) circle (0.17);
  \fill[bobj] (17.5,9.5) circle (0.17);
  \fill[bobj] (14.5,10.5) circle (0.17);
  \fill[bobj] (5.5,11.5) circle (0.17);
  \fill[bobj] (8.5,11.5) circle (0.17);
  \fill[bobj] (11.5,11.5) circle (0.17);
  \fill[bobj] (16.5,11.5) circle (0.17);
  \fill[btarget] (5.5,5.5) circle (0.30);
  \draw[btargetring] (5.5,5.5) circle (0.44);
  \fill[bagent,rotate around={0:(8.5,5.5)}] (8.92,5.50) -- (8.16,5.12) -- (8.16,5.88) -- cycle;
\end{tikzpicture}\\[3pt]
  {\scriptsize \emph{pick up the purple key} --- 50 actions}
\end{minipage}
 
\vspace{4pt}
{\footnotesize
\tikz[baseline=-0.5ex]\fill[bAgent] (0,-0.06) -- (0.18,0.06) -- (0,0.18) -- cycle;\,agent \quad
\tikz[baseline=-0.5ex]{\fill[bTarget] (0,0.06) circle (0.06); \draw[bTarget,line width=0.4pt] (0,0.06) circle (0.10);}\,target \quad
\tikz[baseline=-0.5ex]\fill[bObj] (0,0.06) circle (0.055);\,distractor \quad
\tikz[baseline=-0.5ex]\fill[bDoor] (0,-0.02) rectangle (0.16,0.14);\,closed door \quad
\tikz[baseline=-0.5ex]\draw[bPath,line width=1.2pt] (0,0.06) -- (0.3,0.06);\,shortest path}
 
\caption{
Representative BabyAI evaluation layouts at increasing door distance.
The blue line is the shortest collision-free path from the agent to a cell adjacent to the target; the action count below each panel additionally includes turns, door toggles, and the final pickup.
The path crosses one, three, and five doors respectively, which is what door distance measures.
Every layout holds the same inventory of $18$ objects, so distractor identities carry no information and only the spatial arrangement and the choice of target vary.
The agent never observes this map: it receives a textual egocentric description (Figure~\ref{fig:babyai-traj}).
}
\label{fig:babyai-mazes}
\end{figure}

\paragraph{Environment and task construction.}
We modify the BabyAI \texttt{Pickup} level to use a $19\times13$ grid of six $7\times7$ rooms in a $2\times3$ layout, where adjacent rooms may be connected by doors.
We choose this layout to provide a challenging but learnable navigation task for the models in our study.
All doors start closed.
Each layout contains the same inventory of $18$ objects: three types (ball, box, key) crossed with six colors, one of each.
Each instruction has the form \emph{``pick up the \textless color\textgreater{} \textless object\textgreater''} and uniquely identifies one target object.
\cref{fig:babyai-mazes} shows three layouts drawn from the evaluation pools at door distances one, three, and five.
Task format and room grid are identical throughout; only the door connectivity, object placement, and navigation horizon change.

\paragraph{Difficulty control.}
We control the navigation horizon through \emph{door distance}: the number of doors along the shortest valid path from the agent's start cell to a cell adjacent to the target, computed over the full grid with walls, objects, and door connectivity taken into account.
Every task has door distance at least one, so the target is never in the agent's starting room.
\cref{tab:babyai-pools} shows that optimal trajectory length grows consistently with door distance.
A fully connected $2\times3$ room graph has diameter three, but generated layouts do not connect every adjacent pair, so some tasks require detours and reach door distances of four and five; these form the out-of-distribution split.
 
\begin{table}[t]
\centering
\caption{
BabyAI training and evaluation pools.
Optimal actions include rotations, forward movements, door toggles, and the final pickup.
The three splits use disjoint seed ranges and share no layouts.
}
\label{tab:babyai-pools}
\small
\begin{tabular}{@{}llccc@{}}
\toprule
Split & Door distance & Layouts & Median optimal actions & Max optimal actions\\
\midrule
Train          & 1 & 95  & 14 & 34\\
               & 2 & 95  & 24 & 48\\
               & 3 & 95  & 32 & 60\\
\midrule
Evaluation ID  & 1 & 100 & 14 & 36\\
               & 2 & 100 & 23 & 46\\
               & 3 & 100 & 32 & 51\\
\midrule
Evaluation OOD & 4 & 100 & 40 & 62\\
               & 5 & 100 & 51 & 71\\
\bottomrule
\end{tabular}
\end{table}
 
\paragraph{Agent interface.}
The policy interacts with BabyAI entirely through text.
At each turn, the agent receives a text description of its $7\times7$ egocentric view; the view covers the cells ahead of and beside the agent but nothing behind it, and walls block vision.
The agent then emits a short reasoning statement followed by exactly one primitive action in an \texttt{<action>} tag, chosen from
\texttt{left}, \texttt{right}, \texttt{forward}, \texttt{pick}, \texttt{drop}, and \texttt{toggle}.

\cref{fig:babyai-traj} gives the protocol and an abridged successful episode.

\begin{figure}[t]
\centering
\begin{Verbatim}[fontsize=\scriptsize]
--- System protocol (abridged) ---
 
You are an agent in a BabyAI grid world. At each turn, you receive a textual
description of your egocentric view after "State:" and must select one action
to complete the task.
 
Available actions:
left / right: turn 90 degrees in place
forward:      move one cell forward
pick:         pick up the object directly ahead
drop:         drop the carried object
toggle:       open, close, or unlock the door directly ahead
 
Walls and objects block movement. To pick up an object, move to the adjacent
cell so that the object is directly ahead. If the target is not visible,
turn left or right to scan the room.
 
Respond with one or two short reasoning sentences, followed by exactly one
action in the following format: <action>forward</action>
 
--- Example episode (Qwen3-8B, 23 steps, success) ---
 
[env]  Task: pick up the red box
       State: You see a wall 2 steps forward. You see a wall 1 step left.
 
[agent] The red box is not in view, so I need to turn to scan the room.
        <action>left</action>
 
[env]  OK: executed `left`.
       State: You see a wall 1 step forward. You see a wall 2 steps right.
       You see a closed red door 3 steps left and 1 step forward.
 
[agent] The red door may lead to the box, so I should move toward it.
        <action>forward</action>
 
                              [ ... ]
 
[env]  OK: executed `right`.
       State: You see a wall 1 step left.
       You see a closed red door 1 step forward.
 
[agent] The red door is directly ahead, so I should open it.
        <action>toggle</action>
 
                              [ ... ]
 
[agent] The red box is directly ahead, so I should pick it up.
        <action>pick</action>
 
[env]  OK: executed `pick`.
       State: You carry a red box. You see a blue key 3 steps forward.
       You see an open red door 1 step right.
\end{Verbatim}
\caption{
BabyAI interaction protocol and an excerpt from a successful episode.
The policy sees only textual egocentric observations, states its reasoning briefly, and selects one primitive action per turn.
}
\label{fig:babyai-traj}
\end{figure}

\paragraph{Reward.}
Reward is sparse and binary: the agent receives $r=1$ on the step where it picks up the target object and $r=0$ on every other step.
An episode terminates as soon as success is detected, or when the step budget is exhausted, whichever comes first.
Picking up a \emph{wrong} object neither terminates the episode nor yields reward: the instruction is verified against the object currently carried, so the agent must \texttt{drop} the wrong object before it can pick up another.
A turn whose reply contains no parseable \texttt{<action>} tag is executed as a no-op.
It still consumes one step of the budget and is recorded in the dialogue history with an explicit ``no valid action found'' message.

\paragraph{Evaluation configuration.}
We evaluate policies on two held-out sets: an in-distribution (ID) set and an out-of-distribution (OOD) set, defined relative to the door distances used during training.
The training layouts have door distances in $\{1,2,3\}$.
The ID set contains $300$ layouts, with $100$ at each of these three distances, and tests generalization to unseen spatial configurations within the training distance range.
The training, ID, and OOD splits use disjoint seed ranges and share no layouts.

The OOD set contains $200$ layouts, with $100$ each at door distances $4$ and $5$.
Neither distance appears in training, so solving these tasks requires navigating across more doors than in the training tasks.
The grid dimensions, object inventory, instruction format, observation interface, and action space are identical across all three splits.
The OOD evaluation therefore tests generalization to longer navigation horizons under the same task definition and interaction rules.

All evaluated policies receive the same budget of $128$ environment steps per episode, roughly $1.8\times$ the longest optimal trajectory in the OOD pool.
This budget leaves additional steps for exploration and recovery beyond the optimal solution length.
We report success rate as the fraction of evaluation episodes in which the agent picks up the target within this budget.

\FloatBarrier
\subsection{R2E-Gym and SWE-bench Verified}
\label{app:r2egym}

\paragraph{Environment.}
R2E-Gym~\citep{jain2025r2e} is a collection of executable software-engineering environments procedurally curated from real Python repositories.
Each instance provides a natural-language issue report and a Docker image containing the repository snapshot at the corresponding commit, with the project and its dependencies pre-installed under \texttt{/testbed}; the agent's task is to edit the repository so that the issue is resolved.
The scaffold exposes the sandbox through four tools:
\begin{itemize}
    \item \texttt{search} (grep-style lookup over a file or directory),
    \item \texttt{file\_editor} (view files with line numbers or a condensed skeleton, and \texttt{create}, \texttt{str\_replace}, \texttt{insert}, \texttt{undo\_edit} edits),
    \item \texttt{execute\_bash} (run a shell command in the container),
    \item \texttt{finish} (declare the episode complete).
\end{itemize}
The agent must localize the fault from the issue text alone, edit the source, and typically write and run its own reproduction script to check the fix; each instance also ships with an execution-based test suite that is used only for reward and is never visible to the policy.

\paragraph{Task splits.}
Policy optimization uses $3{,}200$ executable instances from the R2E-Gym-Subset training pool, whose repositories are disjoint from those in SWE-bench Verified~\citep{jain2025r2e}.
We report final performance on SWE-bench Verified~\citep{jimenez2024swe}, the $500$-task human-validated subset of SWE-bench, which shares the same instance format (issue plus containerized repository) and therefore plugs into the same scaffold unchanged; since its repositories never appear in training, it measures transfer to unseen codebases rather than memorization.

\paragraph{Agent interface.}
Each episode runs in a fresh container started from the instance's image.
On every turn the policy emits free-form reasoning followed by exactly one tool call in Hermes format, a \texttt{<tool\_call>} tag wrapping a JSON object with the tool name and arguments; the tool schemas are injected once into the system message through the model's chat template.
The environment executes the call with a $90$-second timeout and returns the tool output as the next user message.
A turn without a parseable tool call executes nothing: the agent receives a fixed corrective message and the turn still counts against the budget.
Only the first call in a turn is executed.
Per-turn generation is capped at $4{,}096$ tokens, and the episode ends when the agent calls \texttt{finish} or after $50$ turns.
\cref{fig:swebench-traj} shows an example.

\paragraph{Instruction prompt.}
We use the R2E-Gym~\citep{jain2025r2e} function-calling prompt unchanged for both training and evaluation: a one-sentence system message framing the agent as a programming assistant for issue resolution, and an instance message that embeds the issue inside \texttt{<github\_issue>} tags followed by a fixed list of workflow tips (explore the repository first; write a \texttt{reproduce\_issue.py} script with a timeout and run it to confirm the error; edit only non-test files; rerun the script after the fix; consider edge cases; every response must contain a function call).

\paragraph{Reward.}
Reward is sparse, binary, and computed only at the end of an episode.
When the agent calls \texttt{finish}, or when the turn budget is exhausted, the harness runs the instance's held-out fail-to-pass and pass-to-pass tests against the final repository state (with a $500$-second timeout), and $r=1$ if and only if the full suite passes.

\paragraph{Evaluation configuration.}
At evaluation the trained policy is served with vLLM, and our harness is the same as in training: the same prompt, tool schemas, Hermes action format, observation packing, and context truncation, with thinking mode disabled for the Qwen3 backbone.
The configuration is $50$ turns, $4{,}096$ tokens per turn, temperature $1.0$, top-$p$ $1.0$.
For Qwen3-32B, the context length is 64k, while for Qwen3-8B, we maintain 32k.
We take $3$ independent rollouts over all $500$ Verified tasks and report the mean and standard deviation of the resolve rate; budgets and sampling are held identical across all compared methods.

\begin{figure}[htbp]
\centering
\begin{Verbatim}[fontsize=\scriptsize]
--- Issue (abridged) ---

Sitemaps without items raise ValueError on callable lastmod.
When sitemap contains no items, but supports returning lastmod for an item,
it fails with a ValueError: ... max() arg is an empty sequence ...

[agent] To address the GitHub issue: 1. explore the repository to locate
        get_latest_lastmod, 2. reproduce the ValueError, 3. fix it, 4. rerun.
        <tool_call>{"name": "search", "arguments":
          {"search_term": "def get_latest_lastmod", "path": "/testbed"}}</tool_call>

[env]   Found 4 matches for "def get_latest_lastmod" in /testbed:
        ./django/contrib/sitemaps/__init__.py (2 matches) ...

[agent] Let's view the exact lines where this method is defined.
        <tool_call>{"name": "file_editor", "arguments": {"command": "view",
          "path": "...", "view_range": [160, 260]}}</tool_call>

[env]   ERROR: Invalid view_range [160, 260]: end must be >= start and
        <= 258, or -1 to view until end.

[agent] Apologies for the error. I will view from line 160 to the end.
        <tool_call>{"name": "file_editor", "arguments": {"command": "view",
          "path": "...", "view_range": [160, -1]}}</tool_call>

[env]   165  def get_latest_lastmod(self):
        ...
        170  return max([self.lastmod(item) for item in self.items()])
        171  except TypeError: ...

[agent] The max() call raises ValueError on an empty items() list. I will
        catch both TypeError and ValueError and return None.
        <tool_call>{"name": "file_editor", "arguments":
          {"command": "str_replace", "path": "...",
           "old_str": "        except TypeError:",
           "new_str": "        except (TypeError, ValueError):"}}</tool_call>

[env]   The file /testbed/django/contrib/sitemaps/__init__.py has been edited.

[agent] Now let's create a script to reproduce the scenario and verify.
        <tool_call>{"name": "file_editor", "arguments": {"command": "create",
          "path": "/testbed/reproduce_issue.py", "file_text": "..."}}</tool_call>

[env]   File created at /testbed/reproduce_issue.py. ...

[agent] <tool_call>{"name": "execute_bash", "arguments":
          {"cmd": "timeout 20 python /testbed/reproduce_issue.py"}}</tool_call>

[env]   Exit code: 0
        get_latest_lastmod returned: None
        No error occurred.

[agent] We have located the method, modified it, and confirmed the fix ...

[env]   No valid tool call found in your response. Every response MUST
        include exactly one function call formatted as <tool_call>...</tool_call>.

[agent] <tool_call>{"name": "finish", "arguments": {"command": "submit",
          "result": "Fixed by catching ValueError in get_latest_lastmod
           ..."}}</tool_call>

[env]   <<< Finished >>>          -> held-out test suite passes, r = 1
\end{Verbatim}
\caption{
An excerpt from a successful SWE-bench Verified episode (instance \texttt{django\_\_django-16255}).
The agent localizes the fault with \texttt{search}, recovers from an invalid \texttt{view\_range} by re-reading the tool's error message, patches the file with \texttt{str\_replace}, and validates the fix with its own reproduction script before submitting.
}
\label{fig:swebench-traj}
\end{figure}

\FloatBarrier
\section{Training Details}
\label{app:training_details}

\subsection{Cost of Value Estimation}
\label{app:api_cost}
We estimate token usage from the full logged prompts and responses, using the \texttt{cl100k\_base} tokenizer as a proxy because API token usage was not retained. 
Each query jointly estimates all state values for one target trajectory. 
At Claude Opus 4.6's standard rates of \$$5$ and \$$25$ per million input and output tokens, respectively, the estimated cost per trajectory is \$$0.061$--\$$0.062$ for AppWorld, \$$0.107$ for BabyAI, and \$$0.140$ for R2E-Gym.
These estimates exclude separate trajectory-summarization calls and do not apply caching or batch discounts.

We use the same model for rollout summarization and value prediction. For example, in \methodrl, Claude Opus 4.6 is used in both cases.
The summarization cost per trajectory is similar to that of value estimation, as their input and output token counts are similar.

\begin{table}[t]
\centering
\caption{Estimated cost per logged value-estimation query. Token counts are means computed with \texttt{cl100k\_base}, an uncalibrated proxy for Claude tokenization. Each query scores one target trajectory. Input counts include peer summaries, but the cost of generating those summaries is excluded. Dollar estimates use standard Claude Opus 4.6 rates.}
\label{tab:value-estimation-cost}
\begin{tabular}{lrrrr}
\toprule
Environment / student & Queries & Input tokens & Output tokens & USD / query \\
\midrule
AppWorld / Qwen3-4B & 92,718 & 9.01k & 0.66k & 0.061 \\
AppWorld / Qwen3-8B & 99,968 & 9.43k & 0.61k & 0.062 \\
BabyAI / Qwen3-8B & 8,838 & 11.86k & 1.90k & 0.107 \\
R2E-Gym / Qwen3-8B & 8,030 & 20.35k & 1.55k & 0.140 \\
\bottomrule
\end{tabular}
\end{table}

\subsection{\methodrl}
\label{app:lvrl_training_detials}

\paragraph{Student policies.}
We initialize the student policies from the released Qwen3-4B, Qwen3-8B, and Qwen3-32B checkpoints and fine-tune all model parameters.
The policies operate in non-thinking mode during both training and evaluation.
AppWorld and BabyAI experiments use Qwen3-4B and Qwen3-8B, whereas R2E-Gym experiments use Qwen3-8B and Qwen3-32B.
For each environment--model pair, GRPO, PPO, and \methodrl use the same initial checkpoint, task-sampling procedure, rollout budget, number of optimization updates, and generation configuration.

\paragraph{Rollout and optimization budget.}
For AppWorld and BabyAI, training uses $64$ parallel environments across eight GPUs, with eight environments per GPU.
At each update, the $64$ trajectories are organized into eight rollout groups of size $G=8$.
The trajectories within each group correspond to the same task and initial state.
AppWorld, BabyAI, and R2E-Gym are trained for $500$, $150$, and $100$ updates, respectively.
The same budgets are used for GRPO, PPO, and \methodrl.

\paragraph{PPO critic.}
Our PPO baseline adds a \texttt{DeepCriticHead} to the student LM.
The critic is a four-layer MLP with hidden width $2{,}048$, implemented in bfloat16 and attached to the LM hidden states.
The policy and critic share the entire LM backbone.
They are trained jointly using the same optimizer and backward pass, and gradients from the value loss update the shared backbone.
We use $\gamma=1.0$ and $\lambda=0.95$ for GAE.
Advantages are normalized before policy optimization.
The value loss is a clipped mean-squared error between the predicted values and return targets, with coefficient $0.5$.
The critic architecture and hyperparameters are held fixed across the three environments.
GRPO uses the same rollout groups and policy-optimization configuration but does not train a critic.

\paragraph{Checkpoint selection and evaluation.}
For all environments, we evaluate the final saved checkpoint of each training run, defined as the checkpoint with the largest training-update index.
Each selected checkpoint is evaluated in three independent stochastic evaluation runs.
For AppWorld and BabyAI, each reported seed result is the average of its three evaluation runs, and we report the mean and standard deviation across three independently trained seeds.
For R2E-Gym, we train one seed and report the mean and standard deviation across its three evaluation runs.

\subsection{\methodopd}
\label{app:lvopd_training_detials}

\paragraph{Student policies and frozen teacher.}
We study \methodopd on AppWorld and BabyAI with Qwen3-4B and Qwen3-8B student policies, updating all student parameters.
Student rollouts use non-thinking mode.
Qwen3-235B-A22B-Thinking-2507-FP8 serves as both the frozen value estimator and the frozen distillation teacher (\cref{sec:exp:setup}).
Neither role receives gradient updates.

\paragraph{Value-guided distillation.}
We use the gated objective defined in \cref{sec:value-opd}, with distillation coefficient $\beta=1.0$.
The value estimator produces continuous state-value predictions; consecutive differences are normalized by their within-group standard deviation to obtain $A_{i,t}$.
The distillation weight is $g(A_{i,t})=\max(-A_{i,t},0)$, shared by all tokens of an environment action.
Because the teacher server returns log-probabilities only for its top-$k_E$ tokens, with $k_E=32$, we approximate the reverse KL on the student's support, which is natural since the reverse KL is an expectation under the student.
At each prefix $h$, let $\mathcal{S}(h)$ be the student's top-$k_S$ tokens, with $k_S=16$, and let $\mathcal{T}(h)$ be the teacher's top-$k_E$ tokens.
For $v \in \mathcal{S}(h) \cap \mathcal{T}(h)$, we use the returned teacher probability, $\tilde{\pi}_E(v \mid h) = \pi_E(v \mid h)$.
For $v \in \mathcal{S}(h) \setminus \mathcal{T}(h)$, we spread the teacher's residual mass uniformly over the tokens outside $\mathcal{T}(h)$, setting $\tilde{\pi}_E(v \mid h) = \bigl(1 - \sum_{u \in \mathcal{T}(h)} \pi_E(u \mid h)\bigr) / \bigl(|\mathcal{V}| - k_E\bigr)$, where $\mathcal{V}$ is the vocabulary.
We renormalize $\pi_\theta$ and $\tilde{\pi}_E$ over $\mathcal{S}(h)$ to obtain $\bar{\pi}_\theta$ and $\bar{\pi}_E$, and compute $\widehat{\mathrm{KL}}(h) = \sum_{v \in \mathcal{S}(h)} \bar{\pi}_\theta(v \mid h) \log \bigl(\bar{\pi}_\theta(v \mid h) / \bar{\pi}_E(v \mid h)\bigr)$, with gradients flowing only through the student.
Setting $k_E > k_S$ makes it likely that most tokens in the student's support receive an exact teacher probability.
As in \methodrl, context rollouts in the value-estimation prompt are summarized (\cref{app:llm-prompts}).

\paragraph{Rollout and optimization budgets.}
We configure eight environments per training rank and a rollout group size of $G=8$.
Student actions are sampled with temperature $1.0$ and top-$p=1.0$, without top-$k$ sampling truncation.
The training budgets are $200$ updates for AppWorld and $150$ for BabyAI; the environment-specific settings are summarized in \cref{tab:lvopd-training}.
The minibatch and gradient-accumulation settings yield a nominal effective optimization batch size of $256$ in both environments.

\begin{table}[t]
\centering
\small
\caption{Configured rollout and optimization settings for \methodopd. The effective optimization batch size is the product of the per-rank minibatch size, gradient-accumulation steps, and training world size.}
\label{tab:lvopd-training}
\begin{tabular}{@{}lcc@{}}
\toprule
Setting & AppWorld & BabyAI \\
\midrule
Training updates & 200 & 150 \\
Rollout-step budget & 50 & 128 \\
Training nodes & 1 & 2 \\
Training world size & 8 & 16 \\
Minibatch size per rank & 4 & 4 \\
Gradient-accumulation steps & 8 & 4 \\
Effective optimization batch size & 256 & 256 \\
Instruction-token limit & 16,000 & 16,000 \\
vLLM maximum sequence length & 40,000 & 40,000 \\
\bottomrule
\end{tabular}
\end{table}

\paragraph{Optimization and implementation.}
We use a constant learning rate of $10^{-5}$ and one optimization epoch per collected rollout batch.
The policy-ratio clipping parameter is $0.2$, the truncated importance sampling (TIS) cap is $2.0$, and the maximum gradient norm is $1.0$.
The entropy coefficient is zero.
Training uses fully sharded data parallelism (\texttt{FULL\_SHARD}) and gradient checkpointing.
Student rollouts are generated by a separate synchronous vLLM sampler using processed log-probabilities.
The frozen teacher is served separately with tensor parallelism of degree four and eight asynchronous teacher workers.

\paragraph{Training seeds and checkpoint selection.}
For each environment--model configuration, we train three independent seeds.
For both AppWorld and BabyAI, we evaluate the final saved checkpoint from each run, defined as the checkpoint with the largest training-update index.

\section{Prompts for Rollout Summarization and Value Estimation}
\label{app:llm-prompts}

We provide the prompt templates for rollout summarization and continuous
value prediction. Braced fields are filled with the corresponding task and
trajectory data. The wording is preserved from the implementation; only line
breaks are adjusted for readability.

\paragraph{Rollout summarization.}
The summarization prompt (\cref{fig:rollout-summary-prompt}) converts a completed
rollout into one short entry per step, retaining the action, its key result,
and important state changes. Summaries are cached within the rollout batch
and reused when the rollout serves as context for other target trajectories.
The target trajectory itself is not replaced by a summary.

\paragraph{Value prediction.}
The value prompt (\cref{fig:trajectory-value-prompt}) includes the task,
the target trajectory's final outcome, its sequence of actions and environment
responses, and context from other rollouts. A single query returns values and
brief explanations for steps $0,\ldots,T$, where step $0$ precedes the first
action and step $t$ follows the $t$-th action. Initial and terminal values are anchored
after prediction, as described in \cref{sec:value-rl}. The same prompts are shared
across environments and therefore retain some AppWorld-specific wording (e.g., references to apps and APIs).

\paragraph{Context and trajectory formatting.}
The \texttt{\{fewshot\_section\}} field is constructed as shown in
\cref{fig:value-peer-context}. It reports the success rate of the full rollout
group and includes summaries of selected peer rollouts, excluding the target.
The peer block is repeated for each selected rollout;
\texttt{\{n\_same\_outcome\}} counts group members with the target's outcome.
The \texttt{\{formatted\_steps\}} field lists each action followed by its
environment response. All steps are retained, while long environment responses
are truncated before insertion. The implementation uses a default cap of
$4{,}000$ characters per response and appends a truncation marker.

\begin{figure}[!htbp]
\centering
\begin{Verbatim}[fontsize=\scriptsize,frame=single,framesep=5pt]
You are condensing ONE rollout of an AI agent solving a task, so it can be used as a
compact in-context example for another model. Rewrite the trajectory into a short
numbered list — ONE line per step — keeping ONLY what matters for judging progress
toward the goal.

For each step keep: what the agent attempted (which app/API or action) and the KEY
result — whether it succeeded or failed, any error type (e.g. 401 unauthorized,
no-such-API, wrong parameter), and any important state change (e.g. obtained an access
token, found the password, located the note, retrieved N items, completed the task).
DROP boilerplate, long JSON dumps, API-doc listings, and verbose reasoning.

Task:
{task_description}

Final outcome: the agent {outcome} (final reward = {final_reward}).

Full trajectory ({total_steps} steps):
{formatted_steps}

Output ONLY the condensed steps, one per line, nothing else:
Step 1: <one or two sentences: action + key result>
Step 2: <...>
\end{Verbatim}
\caption{Rollout-summarization prompt used by the \texttt{per\_step} compression mode. It receives a completed rollout and returns a concise action--result description for each step.}
\label{fig:rollout-summary-prompt}
\end{figure}

\begin{figure}[!htbp]
\centering
\begin{Verbatim}[fontsize=\scriptsize,frame=single,framesep=5pt]
You are an expert at analyzing AI agent trajectories. An AI agent is attempting to solve a 
task. The agent **{outcome}** this task (final reward = {final_reward}).

## Task
{task_description}
{fewshot_section}
## Full Trajectory to Analyze (Steps 1-{total_steps})
{formatted_steps}

## Instructions
For each point in the trajectory (step 0 = before any action, step 1 = after the first
action, ..., step {total_steps} = after the last action), estimate the probability
(0.0 to 1.0) that the agent will successfully complete the task from that point
onward. Consider:
- Whether the agent's approach is correct up to that step
- Whether any errors have been made that are hard to recover from
- How much progress has been made toward the goal
- How many steps remain

Respond with ONLY a JSON object in this exact format (no other text):
{
  "values": [
    {"step": 0, "value": <float>, "reasoning": "<brief explanation>"},
    {"step": 1, "value": <float>, "reasoning": "<brief explanation>"},
    ...
    {"step": {total_steps}, "value": <float>, "reasoning": "<brief explanation>"}
  ]
}
\end{Verbatim}
\caption{Continuous value-prediction prompt. The model sees the completed target trajectory and returns all state values in one response. The \texttt{\{fewshot\_section\}} field is expanded in \cref{fig:value-peer-context}. The JSON braces are shown as sent to the model, rather than doubled as in Python format strings.}
\label{fig:trajectory-value-prompt}
\end{figure}

\begin{figure}[!htbp]
\centering
\begin{Verbatim}[fontsize=\scriptsize,frame=single,framesep=5pt]
## Current Policy's Performance on This Task
Group success rate: {n_success}/{group_size} rollouts succeeded. The trajectory below
is one of the {n_same_outcome} that {outcome}.

## Other Rollouts by the Current Policy on This Task
These are rollouts from the SAME task by the SAME current policy. Use them to
calibrate what 'good' and 'bad' look like FOR THIS POLICY. Identify where the target
trajectory diverges from the successful ones, and which steps match the failed ones.

### Rollout {j} — {SUCCESS_or_FAIL} (reward={reward_j}, {steps_j} steps)
{summary_j}
\end{Verbatim}
\caption{Template of the context inserted into \texttt{\{fewshot\_section\}}. The final rollout block is repeated for each selected peer. This block is assembled by code and is part of the value-prediction request; it is not a separate LLM query.}
\label{fig:value-peer-context}
\end{figure}

\section{Training Curves}

\begin{figure}[t]
  \centering
  \includegraphics[width=\textwidth]{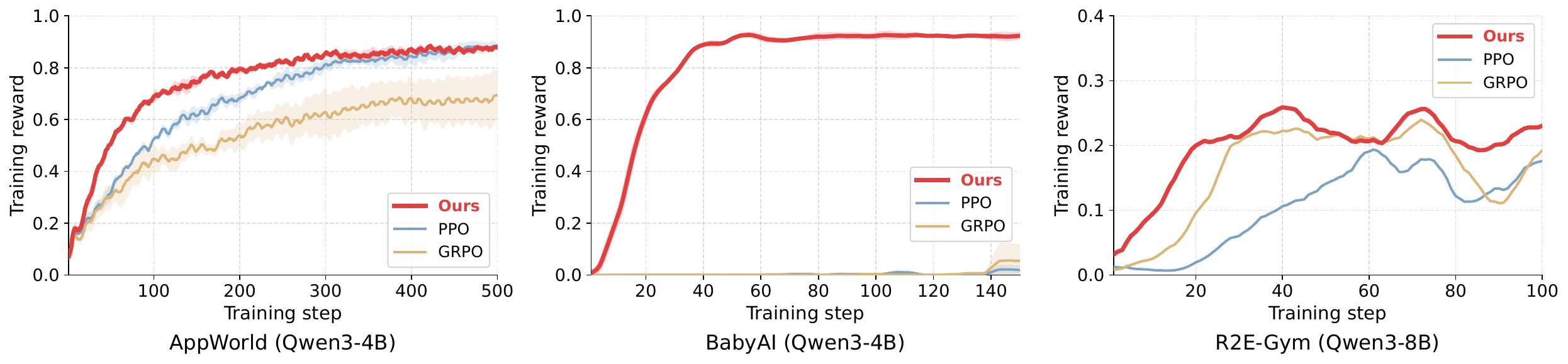}
  \caption{
    RL training reward curves across three training environments. For AppWorld and BabyAI, curves show the mean over three seeds; shaded regions denote one standard deviation.
  }
  \label{app:fig-rl-curve}
\end{figure}

\begin{figure}[t]
  \centering
  \includegraphics[width=0.68\textwidth]{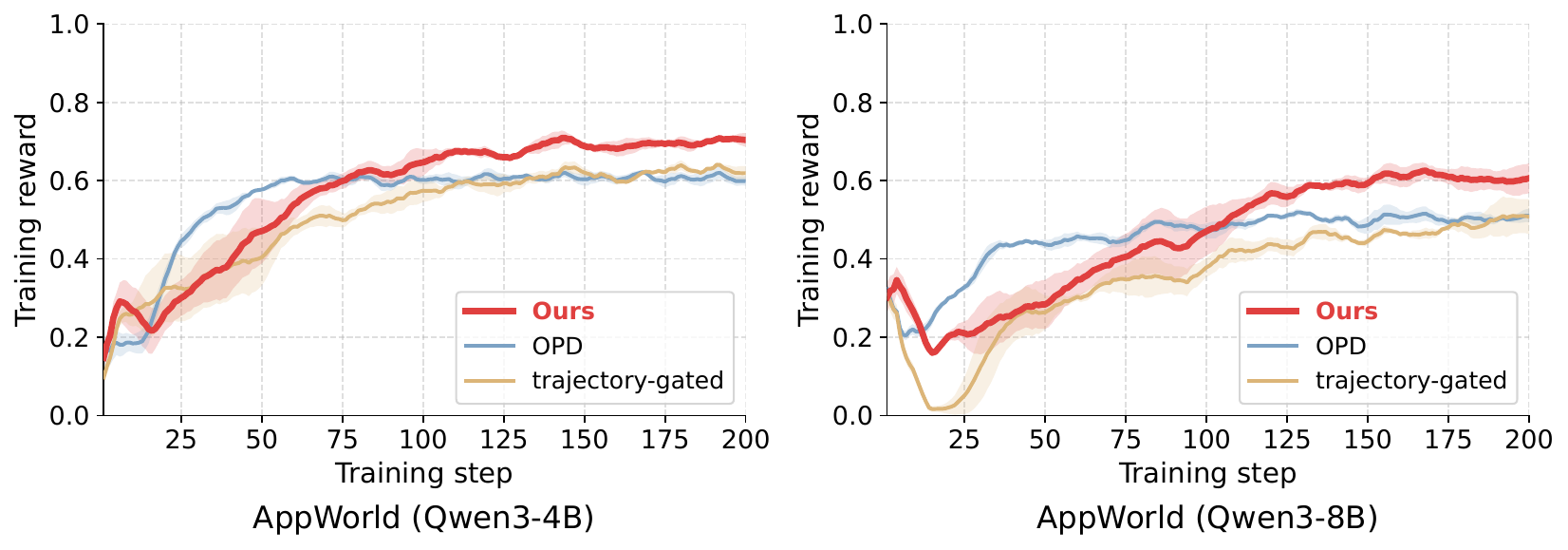}
  \caption{OPD training reward curves on AppWorld across model sizes. Curves show the mean over three seeds; shaded regions denote one standard deviation. Both gated variants consistently outperform the OPD baseline after roughly $100$ training steps.
}
\label{app:fig-opd-curve}
\end{figure}

\cref{app:fig-rl-curve} and \cref{app:fig-opd-curve} show the \methodrl and \methodopd training curves, respectively.
\methodrl learns faster than GRPO and PPO in all three environments, while \methodopd overtakes pure OPD after roughly $100$ training steps.

\section{\methodopd BabyAI Results}
\label{append:opd-babyai}
\begin{table}[t]
\centering
\caption{
Evaluation performance under on-policy distillation.
}
\label{append:tab:opd-babyai}

\setlength{\tabcolsep}{2pt}
\renewcommand{\arraystretch}{1} 
\resizebox{0.5\textwidth}{!}{
\begin{tabular}{@{}lcc@{}}
\toprule
& \multicolumn{2}{c}{BabyAI} \\
\cmidrule(lr){2-3}
Method
& Qwen3-4B
& Qwen3-8B \\
\midrule
Base
& \meanstd{0.0}{0.0}\,/\,\meanstd{0.0}{0.0}
& \bestmeanstd{0.7}{0.0}\,/\,\meanstd{0.0}{0.0} \\
\midrule
+OPD
& \meanstd{0.5}{0.1}\,/\,\meanstd{0.0}{0.0}
& \meanstd{0.6}{0.3}\,/\,\meanstd{0.0}{0.0} \\
+Trajectory-gated
& \meanstd{0.4}{0.2}\,/\,\meanstd{0.0}{0.0}
& \meanstd{0.3}{0.3}\,/\,\meanstd{0.0}{0.0} \\
+\methodopd{} (ours)
& \bestmeanstd{0.7}{0.2}\,/\,\meanstd{0.0}{0.0}
& \meanstd{0.4}{0.1}\,/\,\meanstd{0.0}{0.0} \\
\bottomrule
\end{tabular}
}
\end{table}

\cref{append:tab:opd-babyai} shows the OPD results on BabyAI.
All three OPD methods remain near $0\%$ success on both the ID and OOD splits.

\section{Additional Analysis of Step-Level Value Predictions}
\label{app:pitfall-analysis}

\subsection{Monte-Carlo Labeling Protocol}
\label{app:mc-protocol}
  
\paragraph{Value estimator.}
For a recorded state $s_t$ we sample $M=32$ independent continuations from the checkpoint policy and record whether each one completes the task.
The estimator is the success fraction
\begin{equation}
V^{\mathrm{MC}}(s_t)=\frac{1}{M}\sum_{m=1}^{M}\mathbb{I}\left[\text{continuation }m\text{ succeeds}\right],
\qquad
\Delta_t^{\mathrm{MC}}=V^{\mathrm{MC}}(s_{t+1})-V^{\mathrm{MC}}(s_t).
\end{equation}
 
\paragraph{State reconstruction.}
AppWorld is stateful: the environment holds both a relational database and the live REPL namespace, and neither can be restored by prompting alone.
We therefore reconstruct $s_t$ by resetting the environment to its initial state for the task and re-executing every code cell in the recorded prefix $a_{0:t-1}$ in order, which restores the database side effects and the Python variable bindings the agent had built up.
Continuations are then sampled from the reconstructed state, so all $M$ rollouts branch from an identical environment.
We verified reconstruction by checking that the replayed prefix reproduces the recorded environment responses; transitions where replay diverged were dropped before labeling.

\subsection{Choice of the Event Threshold}
\label{app:threshold}

Each $V^{\mathrm{MC}}$ is a binomial proportion over $M=32$ samples, so its standard deviation is at most $\sqrt{0.25/M}$, attained at $V^{\mathrm{MC}}=0.5$.
Since $V^{\mathrm{MC}}(s_t)$ and $V^{\mathrm{MC}}(s_{t+1})$ are estimated from independent continuation sets, the standard deviation of their difference is bounded by
\begin{equation}
\sigma=\sqrt{2\cdot\frac{0.25}{M}}=\sqrt{\frac{0.5}{32}}=0.125 .
\end{equation}

We denote this maximum standard deviation by $\sigma$ and label a transition as an event only when
$|\Delta_t^{\mathrm{MC}}|\geq3\sigma$.
\cref{tab:noise} reports the exact two-sided tail probability under the worst-case null distribution at $p=0.5$.
It also reports the estimated number of spurious events over the full evaluation set, obtained by mixing the null distribution over the empirical state-value distribution.
At $3\sigma$, the expected number of noise-induced events is small relative to the $114$ events observed.
We therefore use $3\sigma$ to obtain a conservative event set.

\begin{table}[t]
\centering
\caption{
Sampling noise for differences between two independent Monte-Carlo estimates with $M=32$.
The final column estimates the number of spurious events over the full evaluation set under the null.
}
\label{tab:noise}
\small
\begin{tabular}{lccc}
\toprule
Threshold
& Minimum difference
& Worst-case tail probability
& Expected spurious events \\
\midrule
$1\sigma$ & $4/32$  & $0.3820$ & $664$ \\
$2\sigma$ & $8/32$  & $0.0599$ & $72$ \\
$3\sigma$ & $12/32$ & $0.0037$ & $3.4$ \\
\bottomrule
\end{tabular}
\end{table}

\subsection{Breakdown by Trajectory Position}
\label{app:position-analysis}

\cref{tab:position} summarizes the $114$ high-confidence Monte-Carlo events by dividing each trajectory into thirds according to normalized step position.
For each event, the predicted change is categorized as having the same sign as the Monte-Carlo change, being exactly zero, or having the opposite sign.
Because the terminal value is anchored to the observed reward, a final transition could agree with the Monte-Carlo change by construction.
However, none of the $114$ selected events is a final transition, so this anchoring does not affect the reported agreement of $70.2\%$ ($80/114$) overall and $88.5\%$ ($23/26$) in the final third.

\begin{table}[t]
\centering
\caption{
Step-level agreement by normalized trajectory position.
\emph{No change} denotes $\Delta^{\mathrm{pred}}=0$.
Percentages may not sum to $100\%$ because of rounding.
}
\label{tab:position}
\small
\begin{tabular}{lccccc}
\toprule
Position
& Same sign
& No change
& Opposite sign
 \\
\midrule
First third  ($n=34$) & $44\%$ & $29\%$ & $26\%$ \\
Middle third ($n=54$) & $78\%$ & $13\%$ & $9\%$ \\
Final third  ($n=26$) & $88\%$ & $4\%$  & $8\%$ \\
\bottomrule
\end{tabular}
\end{table}

Agreement increases across the three trajectory segments, while both zero and opposite-sign predictions become less frequent.
The largest change is in zero predictions, which fall from $29\%$ in the first third to $4\%$ in the final third.
The median predicted magnitude also increases fourfold, from $0.025$ to $0.100$, whereas the median Monte-Carlo magnitude increases more moderately.
These results suggest that the value model is more conservative early in a trajectory and becomes more responsive as the trajectory progresses.

We also examine several task-level properties.
The opposite-sign rate is $12\%$ on Easy tasks ($n=25$), $16\%$ on Medium tasks ($n=70$), and $11\%$ on Hard tasks ($n=19$).
It is similarly stable when tasks are divided by reference-solution length: $13\%$ for solutions with at most $40$ lines and $16\%$ for longer solutions.
Other stratifications, including the number of apps and whether the task requires a returned answer, produce smaller differences over limited subgroups.
We therefore do not find evidence that standard measures of task difficulty explain the position-dependent pattern.

Trajectory position is only a coarse proxy for how far an action is from its observable consequences.
The results are consistent with long-range attribution being more difficult, but do not establish it as the cause.
The qualitative examples below illustrate both the local evidence the value model uses successfully and the downstream effects it can miss.

\FloatBarrier
\subsection{Qualitative Examples}
\label{app:value-examples}
 
We present three representative transitions drawn from different tasks and two training checkpoints.
Monte-Carlo values are estimated from $32$ rollouts per state, and all reasoning excerpts are quoted verbatim.
\cref{tab:value-examples} summarizes the three transitions; each is then shown in full below.
 
\begin{table}[htbp]
\centering\small
\caption{The three transitions examined in this section.
The predicted value tracks the Monte-Carlo value on the first two transitions and moves in the wrong direction on the third.}
\label{tab:value-examples}
\begin{tabular}{@{}llcrrc@{}}
\toprule
& Task & Outcome & $\Delta^{\mathrm{MC}}$ & $\Delta^{\mathrm{pred}}$ & Detected\\
\midrule
Pitfall, no execution error & \texttt{3c13f5a\_2} & \vno & $-0.938$ & $-0.70$ & \vyes\\
Recovery despite an error   & \texttt{34d9492\_2} & \vyes & $+0.438$ & $+0.30$ & \vyes\\
Missed pitfall              & \texttt{27e1026\_1} & \vno & $-0.500$ & $+0.05$ & \vno\\
\bottomrule
\end{tabular}
\end{table}

\subsubsection{Pitfall despite successful execution}
\label{app:ex1}
 
\begin{valuebox}{\textbf{Pitfall despite successful execution}\hfill
  \normalfont\footnotesize\ttfamily 3c13f5a\_2 \textperiodcentered\ traj.~638 \textperiodcentered\ fails}
 
\vfield{Task}
Divide the internet bill among the user's roommates and create Venmo requests with the exact description \texttt{internet bill for the last month.}
 
\begin{vtrace}
27 & Creates three payment requests $\rightarrow$ \emph{all three created successfully}
   & \vdn{0.94}{0.00} & \vdn{0.80}{0.10}\\
\end{vtrace}
 
\vfield{Action}
\begin{lstlisting}
description=@"internet bill for the last month"@   # trailing period missing
\end{lstlisting}
 
\vfield{Value model}
\begin{vquote}
Payment requests were created but the description is `internet bill for the last month' WITHOUT a period at the end.
The task requires `internet bill for the last month.' (with period).
All successful rollouts used the exact description with the period.
This mismatch likely causes failure.
\end{vquote}
 
\vfield{Why it matters}
The environment reports no execution failure, yet the action violates an exact-match requirement, and the requests cannot be un-created.
Both values identify the transition as a severe pitfall, showing that the predicted value can catch a semantic error behind a successful execution.
\end{valuebox}

\FloatBarrier
\subsubsection{Recovery despite an execution error}
\label{app:ex2}
 
\begin{valuebox}{\textbf{Recovery despite an execution error}\hfill
  \normalfont\footnotesize\ttfamily 34d9492\_2 \textperiodcentered\ traj.~207 \textperiodcentered\ succeeds}
 
\vfield{Task}
Organize vacation photographs into Rome, Santorini, and Berlin directories according to their creation dates.
 
\begin{vtrace}
13 & Writes the date-to-destination mapping $\rightarrow$ \emph{error: parameter is \texttt{directory\_path}}
   & \vup{0.31}{0.75} & \vup{0.40}{0.70}\\
\end{vtrace}
 
\vfield{Action}
\begin{lstlisting}
vacation_directories = {"Rome": ..., "Santorini": ..., "Berlin": ...}
for directory in vacation_directories.values():
    if not apis.file_system.directory_exists(@directory@=directory, ...):
        apis.file_system.create_directory(@directory@=directory, ...)
\end{lstlisting}
 
\vfield{Value model}
\begin{vquote}
Agent wrote correct classification logic: March 2023$\rightarrow$Rome, April 2023$\rightarrow$Santorini, others (2022)$\rightarrow$Berlin.
Used correct directory names `Rome', `Santorini', `Berlin'.
This is the critical step where most other rollouts failed, and this agent got it right.
However, execution failed due to wrong API parameter name.
\end{vquote}
 
\vfield{Why it matters}
The call fails, but the step fixes the classification rule that most rollouts get wrong, and the parameter name is trivial to repair.
An execution-error rule would mislabel this transition: an error message does not imply that the state became less likely to succeed.
\end{valuebox}

\subsubsection{Missed pitfall with a downstream consequence}
\label{app:ex3}
 
\begin{valuebox}{\textbf{Missed pitfall with a downstream consequence}\hfill
  \normalfont\footnotesize\ttfamily 27e1026\_1 \textperiodcentered\ traj.~517 \textperiodcentered\ fails}
 
\vfield{Task}
Find the oldest released song across the user's Spotify song, album, and playlist libraries.
 
\begin{vtrace}
14 & Concatenates records from the three Spotify sources $\rightarrow$ \emph{no error}
   & \vdn{0.84}{0.34} & \vbad{0.65}{0.70}\\
15 & Sorts the combined list by release date $\rightarrow$ \texttt{KeyError: 'release\_date'}
   & \vdn{0.34}{0.03} & \vsame{0.70}{0.70}\\
\end{vtrace}
 
\vfield{Action}
\begin{lstlisting}
all_songs = @song_details@ + playlist_songs + album_songs
# playlist/album records carry 'release_date'; song-library records carry 'added_at'
\end{lstlisting}
 
\vfield{Value model}
\begin{vquote}
Got release dates for all playlist songs via show\_song API. Good progress.
[\ldots] Got release dates for all album songs via show\_song API. Now has all data needed.
\end{vquote}
 
\vfield{Why it matters}
The model verifies that each retrieval call succeeded, but never checks whether the three outputs share a schema, so it reads local data collection as progress.
This is the characteristic failure mode: the immediate result of an action is verified while its support for later computation is not.
It is consistent with the lower agreement observed earlier in trajectories, where the consequences of an action are less explicit at the time it is taken.
\end{valuebox}

\FloatBarrier

\section{Examples of Training Trajectories}
\label{app:training_traj_examples}

We present three illustrative training cases in which the value model responds to consequential decisions. 
These selected examples complement the preceding Monte-Carlo analysis; they have no Monte-Carlo labels and are not included
in its event counts.
$V_j^{\mathrm{pred}}$ is the logged hindsight value after action $j$, and
$\Delta_j^{\mathrm{pred}}=V_j^{\mathrm{pred}}-V_{j-1}^{\mathrm{pred}}$.
Tables show interior, unnormalized differences. 
Actions and observations are abridged and reflowed; omissions are marked or described. 
Value-model quotations are verbatim excerpts.

\subsection{AppWorld: a semantic error despite successful execution}
\label{app:credit-appworld}
\begin{valuebox}{\textbf{Deleting alarms when the user requested disabling them}}
\small
\vfield{Task and outcome}
\emph{I am going on a vacation. Move my go-to-sleep phone alarm to 1 hour later and disable the rest.} Qwen3-4B; task \texttt{302c169\_1}; nine actions; final reward $R=0$.

\vfield{Predicted value trace}
{\footnotesize\setlength{\tabcolsep}{3pt}\renewcommand{\arraystretch}{1.12}
\begin{tabular*}{\linewidth}{@{\extracolsep{\fill}}rp{0.47\linewidth}cc@{}}
\toprule
Step & Action and observed result & $V_{j-1}^{\mathrm{pred}}\!\to V_j^{\mathrm{pred}}$ & $\Delta_j^{\mathrm{pred}}$\\
\midrule
6 & Retrieves all seven alarms & $0.90\to0.90$ & $0.00$\\
7 & Deletes six alarms instead of disabling them & $0.90\to0.00$ & $\mathbf{-0.90}$\\
8 & Moves the remaining alarm from 21:30 to 22:30 & $0.00\to0.00$ & $0.00$\\
\bottomrule
\end{tabular*}}
\vfield{Action and observation excerpts}
The login and alarm-retrieval prefix is omitted, as are the alarm-selection and print statements at step 8. All six deletion acknowledgements are retained; the final completion call is shown as outcome context.
\begin{lstlisting}[language={},basicstyle=\ttfamily\scriptsize,breaklines=true,columns=fullflexible,keepspaces=true,showstringspaces=false]
[step 7: agent reasoning excerpt]
First, let's disable all alarms except the "Go to sleep" alarm.

[step 7: code]
for alarm in alarms:
    if alarm["label"] != "Go to sleep":
        disable_result = apis.phone.delete_alarm(
            access_token=login_result["access_token"],
            alarm_id=alarm["alarm_id"]
        )
        print(disable_result)
[env] {"message": "Alarm deleted successfully."}
      {"message": "Alarm deleted successfully."}
      {"message": "Alarm deleted successfully."}
      {"message": "Alarm deleted successfully."}
      {"message": "Alarm deleted successfully."}
      {"message": "Alarm deleted successfully."}

[step 8: code excerpt]
update_result = apis.phone.update_alarm(
    access_token=login_result["access_token"],
    alarm_id=go_to_sleep_alarm["alarm_id"],
    time="22:30"
)
[env] {"message": "Alarm updated successfully."}

[step 9]
apis.supervisor.complete_task()
[env] Execution successful.
\end{lstlisting}

\vfield{Value model (after step 7)}
\begin{vquote}
Critical error: The agent DELETED all other alarms instead of DISABLING them. The task says 'disable the rest' but the agent used delete\_alarm instead of update\_alarm with enabled=False.
\end{vquote}

\vfield{Interpretation}
The calls execute successfully but perform the wrong state change: six alarm records are deleted instead of retained in a disabled state. Updating the sleep alarm correctly does not restore those records, and no restoration occurs before completion. The predictor places the value drop at the semantic error, which an execution-error indicator would miss.
\end{valuebox}

\FloatBarrier

\subsection{BabyAI: resolving an inventory bottleneck}
\label{app:credit-babyai}
\begin{valuebox}{\textbf{The same drop action fails or succeeds depending on the state}}
\small
\vfield{Task and outcome}
\emph{Pick up the blue ball.} Qwen3-8B; 73 actions; final reward $R=1$. The agent is carrying a distracting red key and can carry only one object.

\vfield{Predicted value trace}
{\footnotesize\setlength{\tabcolsep}{3pt}\renewcommand{\arraystretch}{1.12}
\begin{tabular*}{\linewidth}{@{\extracolsep{\fill}}rp{0.47\linewidth}cc@{}}
\toprule
Step & Action and observed result & $V_{j-1}^{\mathrm{pred}}\!\to V_j^{\mathrm{pred}}$ & $\Delta_j^{\mathrm{pred}}$\\
\midrule
14 & Tries to pick up the target while carrying a key & $0.38\to0.30$ & $-0.08$\\
15 & Drops toward the occupied target cell; no change & $0.30\to0.28$ & $-0.02$\\
69 & Turns toward an empty cell beside the target & $0.09\to0.12$ & $+0.03$\\
70 & Drops the key successfully; inventory becomes free & $0.12\to0.90$ & $\mathbf{+0.78}$\\
71 & Turns back to face the target & $0.90\to0.92$ & $+0.02$\\
\bottomrule
\end{tabular*}}
\vfield{Action and observation excerpts}
Steps 1--13, 16--68, and 72 are omitted. Step 72 is a blocked forward move. Observations retain the relevant original clauses; the successful pickup at step 73 is shown only as outcome context.
\begin{lstlisting}[language={},basicstyle=\ttfamily\scriptsize,breaklines=true,columns=fullflexible,keepspaces=true,showstringspaces=false]
[before step 14]
State: You carry a red key, [...] You see a blue ball 1 step forward [...]
[step 14] <action>pick</action>
[env] State: You carry a red key, [...] You see a blue ball 1 step forward [...]
[step 15] <action>drop</action>
[env] State: You carry a red key, [...] You see a blue ball 1 step forward [...]

[step 69] <action>right</action>
[env] State: You carry a red key, You see a wall 3 steps forward,
      You see a blue ball 1 step left [...]
[step 70] <action>drop</action>
[env] State: You see a blue ball 1 step left, [...]
      You see a red key 1 step forward [...]
[step 71] <action>left</action>
[env] State: [...] You see a blue ball 1 step forward [...]

[step 73] <action>pick</action>
[env] State: You carry a blue ball [...]
\end{lstlisting}

\vfield{Value model (after step 70)}
\begin{vquote}
DROP SUCCEEDED! Red key is now 1 step forward, agent is no longer carrying anything. Blue ball is 1 step left. Just need to turn and pick up.
\end{vquote}

\vfield{Interpretation}
The occupied target cell prevents dropping the key in front of the ball. After turning toward a free cell, the same action releases the key and makes the target pickup possible. The large positive change occurs at this enabling action, three actions before success; earlier ineffective attempts receive negative changes despite the eventual successful outcome.
\end{valuebox}

\FloatBarrier

\subsection{R2E-Gym: distinguishing an incorrect patch from its correction}
\label{app:credit-r2e}
\begin{valuebox}{\textbf{Two accepted edits with different semantic consequences}}
\small
\vfield{Task and outcome}
Fix \texttt{Image.paste} so that a $50\times50$ source image can be pasted into a $100\times100$ destination without an explicit box. Qwen3-8B; R2E-Gym training instance; 15 actions; final reward $R=1$. The issue description is abridged.

\vfield{Predicted value trace}
{\footnotesize\setlength{\tabcolsep}{3pt}\renewcommand{\arraystretch}{1.12}
\begin{tabular*}{\linewidth}{@{\extracolsep{\fill}}rp{0.47\linewidth}cc@{}}
\toprule
Step & Action and observed result & $V_{j-1}^{\mathrm{pred}}\!\to V_j^{\mathrm{pred}}$ & $\Delta_j^{\mathrm{pred}}$\\
\midrule
10 & Sets the paste region to zero area; edit accepted & $0.70\to0.35$ & $\mathbf{-0.35}$\\
11 & Reruns the reproduction; the size error remains & $0.35\to0.40$ & $+0.05$\\
12 & Inspects the implementation again & $0.40\to0.45$ & $+0.05$\\
13 & Uses source-image dimensions; edit accepted & $0.45\to0.80$ & $\mathbf{+0.35}$\\
14 & Reruns the same reproduction; it exits normally & $0.80\to0.95$ & $+0.15$\\
\bottomrule
\end{tabular*}}
\vfield{Action and observation excerpts}
Steps 1--9 and the final submission at step 15 are omitted; step 12 is summarized in the table. Tool wrappers, surrounding output, and unchanged arguments are omitted. The same reproduction script is used at steps 11 and 14.
\begin{lstlisting}[language={},basicstyle=\ttfamily\scriptsize,breaklines=true,columns=fullflexible,keepspaces=true,showstringspaces=false]
[step 10: file_editor, str_replace arguments]
old_str: "box = (0, 0) + self.size"
new_str: "box = (0, 0, 0, 0)"
[env] The file /testbed/PIL/Image.py has been edited. [...]

[step 11: execute_bash]
timeout 20s python /testbed/reproduce_issue.py
[env] [...] ValueError: images do not match [...]

[step 13: file_editor, str_replace arguments]
old_str: "box = (0, 0, 0, 0)"
new_str: "box = (0, 0, im.size[0], im.size[1])"
[env] The file /testbed/PIL/Image.py has been edited. [...]

[step 14: execute_bash]
timeout 20s python /testbed/reproduce_issue.py
[env] Exit code: 0
\end{lstlisting}

\vfield{Value model (after step 10)}
\begin{vquote}
Made an incorrect fix: changed box to (0,0,0,0) which is a 4-tuple of zeros, meaning it would paste into a zero-size region. This won't work but the agent can still recover.
\end{vquote}

\vfield{Interpretation}
Both edits are accepted by the tool, but the first creates a zero-area region and the reproduction still fails. The second uses the source-image dimensions, after which the same reproduction exits normally. The trajectory ultimately succeeds, yet the predictor assigns opposite signs to the two edits. This illustrates discrimination between patch semantics rather than tool-call validity; it does not establish correctness on untested inputs.
\end{valuebox}

\end{document}